\def\year{2022}\relax
\documentclass[letterpaper]{article} % DO NOT CHANGE THIS
\usepackage{aaai22}  % DO NOT CHANGE THIS
\usepackage{times}  % DO NOT CHANGE THIS
\usepackage{helvet}  % DO NOT CHANGE THIS
\usepackage{courier}  % DO NOT CHANGE THIS
\usepackage[hyphens]{url}  % DO NOT CHANGE THIS
\usepackage{graphicx} % DO NOT CHANGE THIS
\usepackage{natbib}  % DO NOT CHANGE THIS AND DO NOT ADD ANY OPTIONS TO IT
\usepackage{caption} % DO NOT CHANGE THIS AND DO NOT ADD ANY OPTIONS TO IT
\DeclareCaptionStyle{ruled}{labelfont=normalfont,labelsep=colon,strut=off} % DO NOT CHANGE THIS
\usepackage{algorithm}[1]
\usepackage{algorithmic}[1]
\usepackage[algo2e]{algorithm2e} 

\usepackage{newfloat}
\usepackage{listings}
\floatstyle{ruled}
\newfloat{listing}{tb}{lst}{}
\floatname{listing}{Listing}

\author{
    Akshita Jha and %\textsuperscript{\rm 1} and
    Chandan K. Reddy \\ %\textsuperscript{\rm 1}\\
}
\affiliations{
    Department of Computer Science, Virginia Tech, Arlington VA - 22203.\\
    akshitajha@vt.edu, reddy@cs.vt.edu
}

\usepackage{amsmath}
\usepackage{multirow}
\usepackage{graphicx}

\usepackage{float}
\usepackage{subcaption}
\usepackage{wrapfig}

\lstdefinestyle{C} {language=C}
\lstnewenvironment{C}{
\lstset{
    language=C++,
    basicstyle=\linespread{0.8}\ttfamily\tiny,
    aboveskip=-0.3em,
    belowskip=-1.5em,
    breaklines=true,
    escapeinside={<@}{@>},
    keywordstyle=\color{blue}\bfseries,
    commentstyle=\color{green},
    stringstyle=\ttfamily\color{red!50!brown},
    showstringspaces=false
}
\lstset{literate=%
   *{0}{{{\color{red!20!violet}0}}}1
    {1}{{{\color{red!20!violet}1}}}1
    {2}{{{\color{red!20!violet}2}}}1
    {3}{{{\color{red!20!violet}3}}}1
    {4}{{{\color{red!20!violet}4}}}1
    {5}{{{\color{red!20!violet}5}}}1
    {6}{{{\color{red!20!violet}6}}}1
    {7}{{{\color{red!20!violet}7}}}1
    {8}{{{\color{red!20!violet}8}}}1
    {9}{{{\color{red!20!violet}9}}}1
}
}{}
\newcommand{\model}{CodeAttack}
\newcommand{\tb}{\textbf}

\usepackage{xcolor}
\usepackage{color,soul}
\definecolor{ForestGreen}{RGB}{34,139,34}

\title{CodeAttack: Code-Based Adversarial Attacks for Pre-trained Programming Language Models}

\begin{document}
\maketitle
\begin{abstract}
Pre-trained programming language (PL) models (such as CodeT5, CodeBERT, GraphCodeBERT, etc.,) have the potential to automate software engineering tasks involving code understanding and code generation. However, these models operate in the natural channel of code, \textit{i.e.}, they are primarily concerned with the human understanding of the code. They are not robust to changes in the input and thus, are potentially susceptible to adversarial attacks in the natural channel. We propose, {\textbf{\model}}, a simple yet effective black-box attack model that uses code structure to generate effective, efficient, and imperceptible adversarial code samples and demonstrates the vulnerabilities of the state-of-the-art PL models to code-specific adversarial attacks. We evaluate the transferability of {\model} on several code-code (translation and repair) and code-NL (summarization) tasks across different programming languages. {\model} outperforms state-of-the-art adversarial NLP attack models to achieve the best overall drop in performance while being more efficient, imperceptible, consistent, and fluent. The code can be found at {https://github.com/reddy-lab-code-research/CodeAttack}.

\end{abstract}

\section*{Introduction}
% What is adversarial attack. Why is it important in general. Mention that nothing has been done in the code domain. 

{There has been a recent surge in the development of general purpose programming language (PL) models \cite{ahmad2021unified, feng2020codebert, guo2020graphcodebert, tipirneni2022structcoder, wang2021codet5}. They can capture the relationship between natural language and source code, and potentially automate software engineering development tasks involving code understanding (clone detection, defect detection) and code generation (code-code translation, code-code refinement, code-NL summarization).}
However, the data-driven pre-training of the above models on massive amounts of code data constraints them to primarily operate in the `natural channel' of code \cite{chakraborty2022natgen, hindle2016naturalness, zhang2022diet}. \textit{This `natural channel' focuses on conveying information to humans through code comments, meaningful variable names, and function names} 
% rather than on code compilation and execution 
\cite{casalnuovo2020theory}. In such a scenario, the robustness and vulnerabilities of the pre-trained models need careful investigation. In this work, we leverage the code structure to \textit{generate adversarial samples in the natural channel of code} and demonstrate the vulnerability of the state-of-the-art programming language models to adversarial attacks.

\begin{figure}
     \includegraphics[width=0.48\textwidth]{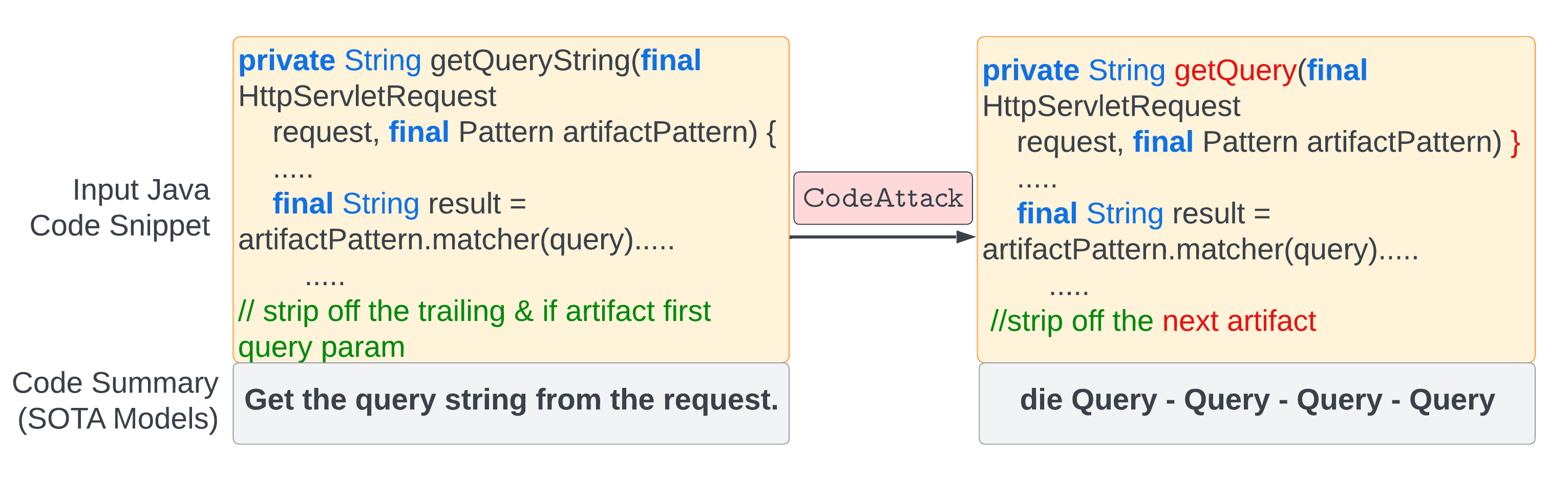}
    \caption{
    {\model} makes a small modification to the input code snippet (in {red}) which causes significant changes to the code summary obtained from the SOTA pre-trained programming language models. Keywords are highlighted in {blue} and comments in {green}.}
    \label{fig:intro}
\end{figure}

% Why is adversarial attack important in the code domain.
Adversarial attacks are characterized by imperceptible changes in the input that result in incorrect predictions from a machine learning model. For pre-trained PL models operating in the natural channel, such attacks are important for two primary reasons: (i) \textit{Exposing system vulnerabilities} and \textit{evaluating model robustness}: A small change in the input programming language (akin to a typo in the NL scenario) can trigger the code summarization model to generate gibberish natural language code summary (Figure~\ref{fig:intro}), and (ii) \textit{Model interpretability}: Adversarial samples can be used to inspect the tokens pre-trained PL models attend to.

%  What are the challenges of adversarial attack in the code domain. Why won't existing NLP adversarial attack methods work in the code domain.
A successful adversarial attack in the natural channel for code should have the following properties: (i) \textit{Minimal perturbations}: Akin to spelling mistakes or synonym replacement in NL that mislead neural models with imperceptible changes, (ii) \textit{Code Consistency}: Perturbed code is consistent with the original input and follows the same coding style as the original code, and (iii) \textit{Code fluency}: Does not alter the user-level code understanding of the original code. The current natural language adversarial attack models fall short on all three fronts. Hence, we propose {\textbf{\model}} -- a simple yet effective black-box attack model for generating adversarial samples in the natural channel for any input code snippet, irrespective of the programming language. 

%  What do we propose?
{\model} operates in a realistic scenario, where the adversary does \textbf{not} have access to model parameters but only to the test queries and the model prediction. {\model} uses a pre-trained masked CodeBERT model \cite{feng2020codebert} as the adversarial code generator to generate imperceptible and effective adversarial examples by leveraging the code structure.
%  What are our main contributions.
Our primary contributions are as follows:
\begin{itemize}
    \item To the best of our knowledge, our work is the first one to detect the vulnerabilities of pre-trained programming language models to adversarial attacks in the natural channel of code. We propose a simple yet effective realistic black-box attack method, {\model}, that generates adversarial samples for a code snippet irrespective of the input programming language.
    \item We design a general purpose black-box attack method for sequence-to-sequence PL models that is transferable across different downstream tasks like code translation, repair, and summarization. The input language agnostic nature of our method also makes it extensible to sequence-to-sequence tasks in other domains.
    \item We demonstrate the effectiveness of {\model} over existing NLP adversarial models through an extensive empirical evaluation. {\model} outperforms the natural language baselines when considering both the attack quality and its efficacy.
\end{itemize}

\section*{Background and Related Work}
\vspace{5pt}
\noindent\textbf{Dual Channel of Source Code.} \citet{casalnuovo2020theory} proposed a dual channel view of code: (i) formal, and (ii) natural. The formal channel is precise and used for code execution by compilers and interpreters. The natural language channel, on the other hand, is for human comprehension and is noisy. It relies on code comments, variable names, function names, etc., to ease human understanding. The state-of-the-art PL models operate primarily in the natural channel of code \cite{zhang2022diet} and therefore, we generate \textit{adversarial samples by making use of this natural channel}.

\vspace{5pt}
\noindent\textbf{Adversarial Attacks in NLP.} BERT-Attack \cite{li2020bert} and BAE \cite{garg2020bae} use BERT for attacking vulnerable words. TextFooler \cite{jin2020bert} and PWWS \cite{ren-etal-2019-generating} use synonyms and part-of-speech (POS) tagging to replace important tokens. Deepwordbug \cite{gao2018black} and TextBugger \cite{li2019textbugger} use character insertion, deletion, and replacement strategy for attacks whereas \citet{hsieh2019robustness} and \citet{yang2020greedy} use a greedy search and replacement strategy. \citet{alzantot2018generating} use genetic algorithm and \citet{ebrahimi2018hotflip}, \citet{papernot2016technical}, and \citet{pruthi2019combating} use model gradients for finding subsititutes. None of these methods have been designed specifically for programming languages, which is more structured than natural language. 

\vspace{5pt}
\noindent\textbf{Adversarial Attacks for PL.} \citet{zhang2020generating} generate adversarial examples by renaming identifiers using Metropolis-Hastings sampling \cite{metropolis1953equation}. \citet{yang2022natural} improve on that by using greedy and genetic algorithm. \citet{yefet2020adversarial} use gradient based exploration; whereas \citet{applis2021assessing} and \cite{ramakrishnan2020semantic} propose metamorphic transformations for attacks. The above models focus on classification tasks like defect detection and clone detection. Although some works do focus on adversarial examples for code summarization \cite{ramakrishnan2020semantic, zhou2021adversarial}, they do not do so in the natural channel. They also do not test the transferability to different tasks, PL models, and different programming languages. Our model, {\model}, assumes black-box access to the state-of-the-art PL models for generating adversarial attacks for code generation tasks like code translation, code repair, and code summarization using a constrained code-specific greedy algorithm to find meaningful substitutes for vulnerable tokens, irrespective of the input programming language.

\section*{{\model}}
We describe the capabilities, knowledge, and the goal of the proposed model, and provide details on how it detects vulnerabilities in the state-of-the-art pre-trained PL models.

\subsection{Threat Model}
\paragraph{Adversary's Capabilities.} The adversary is capable of perturbing the test queries given as input to a pre-trained PL model to generate adversarial samples. We follow the existing literature for generating natural language adversarial examples and allow for two types of perturbations for the input code sequence in the natural channel: (i) character-level perturbations, and (ii) token-level perturbations. The adversary is allowed to perturb only a certain number of tokens/characters and must ensure a high similarity between the original code and the perturbed code. Formally, for a given input code sequence $\mathcal{X} \in X$, where $X$ is the input space, a valid adversarial code example $\mathcal{X}_{adv}$ satisfies the requirements:
\begin{equation}
\mathcal{X} \ne \mathcal{X}_{adv}
\label{eq:x_xadv}
\end{equation}
\begin{equation}
\mathcal{X}_{adv} \leftarrow \mathcal{X} + \delta; \qquad \text{s.t.}~||\delta|| < \theta   
\label{eq:xadv_delta}
\end{equation}
\begin{equation}
\text{Sim}(\mathcal{X}_{adv}, \mathcal{X}) \ge \epsilon    
\label{eq:sim}
\end{equation}
where $\theta$ is the maximum allowed perturbation; $\text{Sim}(\cdot)$ is a similarity function;
and $\epsilon$ is the similarity threshold. 
% We describe the perturbation constraints and the similarity functions in more detail later in Section~\ref{sec:substitutes}.

\paragraph{Adversary's Knowledge.} We assume standard black-box access to realistically assess the vulnerabilities and robustness of existing pre-trained PL models. The adversary does \textit{not} have access to the model parameters, model architecture, model gradients, training data, or the loss function. It can only query the pre-trained PL model with input sequences and get their corresponding output probabilities. This is more practical than a white-box scenario where the attacker assumes access to all the above.

\paragraph{Adversary's Goal.} Given an input code sequence as query, the adversary's goal is to degrade the quality of the generated output sequence through imperceptibly modifying the query in the natural channel of code. The generated output sequence can either be a code snippet (code translation, code repair) or natural language text (code summarization). Formally, given a pre-trained PL model $F: X \to Y$, where $X$ is the input space, and $Y$ is the output space, the goal of the adversary is to generate an adversarial sample $\mathcal{X}_{adv}$ for an input sequence $\mathcal{X}$ \textit{s.t.} 
\begin{equation}
F(\mathcal{X}_{adv}) \ne F(\mathcal{X})  
\label{eq:y_ydash}
\end{equation}
\begin{equation}
Q(F(\mathcal{X})) - Q(F(\mathcal{X}_{adv})) \ge \phi
\label{eq:qual}
\end{equation}
where $Q(\cdot)$ measures the quality of the generated output and $\phi$ is the specified drop in quality. This is in addition to the constraints applied on $\mathcal{X}_{adv}$ earlier.
We formulate our final problem of generating adversarial samples as follows:
\begin{equation}
\Delta_{atk} = \text{argmax}_{\delta}\;  [Q(F(\mathcal{X})) - Q(F(\mathcal{X}_{adv}))]
\label{eq:optimization}
\end{equation}
In the above objective function, $\mathcal{X}_{adv}$ is a minimally perturbed adversary subject to constraints on the perturbations $\delta$ (Eqs.\ref{eq:x_xadv}-\ref{eq:qual}). {\model} searches for a perturbation $\Delta_{atk}$ to maximize the difference in the quality $Q(\cdot)$ of the output sequence generated from the original input code snippet $\mathcal{X}$ and that by the perturbed code snippet $\mathcal{X}_{adv}$.

\subsection{Attack Methodology}

There are two primary steps: (i) Finding the most vulnerable tokens, and (ii) Substituting these vulnerable tokens (subject to code-specific constraints), to generate adversarial samples in the natural channel of code.

\subsubsection{Finding Vulnerable Tokens}\label{sec:vuln}
CodeBERT gives more attention to keywords and identifiers while making predictions \cite{zhang2022diet}. We leverage this information and hypothesize that certain input tokens contribute more towards the final prediction than others. `Attacking' these highly influential or highly vulnerable tokens increases the probability of altering the model predictions more significantly as opposed to attacking non-vulnerable tokens. Under a black-box setting, the model gradients are unavailable and the adversary only has access to the output logits of the pre-trained PL model. We define `vulnerable tokens' as tokens having a high influence on the output logits of the model. Let $F$ be an encoder-decoder pre-trained PL model. The given input sequence is denoted by $\mathcal{X}=[x_1,..,x_i,...,x_m]$, where $\{x_i\}_{1}^{m}$ are the input tokens. The output is a sequence of vectors:
    $\mathcal{O} = F(\mathcal{X}) = [o_1,...,o_n]$; 
    $y_t = \text{argmax}(o_t)$;
where $\{o_t\}_{1}^{n}$ is the output logit for the correct output token $y_t$ for the time step $t$. Without loss of generality, we can also assume the output sequence $\mathcal{Y} = F(\mathcal{X}) = [y_i,...,y_l]$. $\mathcal{Y}$ can either be a sequence of code or natural language tokens. 

To find the vulnerable input tokens, we replace a token $x_i$ with $\textnormal{[MASK]}$ such that $\mathcal{X}_{\setminus x_i} = [x_1,.,x_{i-1},\textnormal{[MASK]},x_{i+1},.,x_m]$ and get its output logits. The output vectors are now
$\mathcal{O}_{\setminus x_i} = F(\mathcal{X}_{\setminus x_i}) = [o'_1,...,o'_q]$
where $\{o'_t\}_{1}^{q}$ is the new output logit for the correct prediction $\mathcal{Y}$. The influence score for the token $x_i$ is as follows:
\begin{equation}
    I_{x_i} = \sum_{t=1}^{n}{o_t} - \sum_{t=1}^{q}{o'_{t}}
\label{eq:infl}
\end{equation}
We rank all the tokens according to their influence score $I_{x_i}$ in descending order to find the most vulnerable tokens $V$. We select the top-$k$ tokens to limit the number of perturbations and attack them iteratively either by replacing them or by inserting/deleting a character around them. 
% We explain this in detail below.

\subsubsection{Substituting Vulnerable Tokens}\label{sec:substitutes}
We adopt greedy search using a masked programming language model, subject to code-specific constraints, to find substitutes $S$ for vulnerable tokens $V$ such that they are minimally perturbed and have the maximal probability of incorrect prediction. 

\vspace{5pt}
\noindent\textit{\textbf{Search Method.}} In a given input sequence, we mask a vulnerable token $v_i$ and use the masked PL model to predict a meaningful contextualized token in its place. We use the top-$k$ predictions for each of the masked vulnerable tokens as our initial search space. Let $\mathcal{M}$ denote a masked PL model. Given an input sequence $\mathcal{X} = [x_1,..,v_i,..,x_m]$, where $v_i$ is a vulnerable token, $\mathcal{M}$ uses WordPiece algorithm \cite{wu2016google} for tokenization that breaks uncommon words into sub-words resulting in $\mathcal{H} = [h_1,h_2,..,h_q]$. We align and mask all the corresponding sub-words for $v_i$, and combine the predictions to get the top-$k$ substitutes $S' = \mathcal{M}(H)$ for the vulnerable token $v_i$. This initial search space $S'$ consists of $l$ possible substitutes for a vulnerable token $v_i$. We then filter out substitute tokens to ensure minimal perturbation, code consistency, and code fluency of the generated adversarial samples, subject to code-specific constraints.

 \begin{table}[]
  \small
     \centering
     \begin{tabular}{r|p{5.3cm}}
     \hline
     \rule{0pt}{2.5ex}\tb{Token Class} & \tb{Description} \\[0.03cm]
     \hline
     \rule{0pt}{2.5ex}
     {Keywords} & Reserved word \\
     {Identifiers} & Variable, Class Name, Method name\\
     {Operators} & Brackets (\{\},(),[]), Symbols (+,*,/,-,\%,;,.)\\ 
     {Arguments} & Integer, Floating point, String, Character \\ [0.07cm]
    \hline
 \end{tabular}
     \caption{Token class and their description.}
     \label{tab:token_class}
 \end{table}
 
\vspace{5pt}
\noindent\textit{\textbf{Code-Specific Constraints.}} Since the tokens generated from a masked PL model may not be meaningful individual code tokens, we further use a CodeNet tokenizer \cite{puri2021project} to break a token into its corresponding code tokens. The code tokens are tokenized into four primary code token classes (Table \ref{tab:token_class}). If $s_i$ is the substitute for the vulnerable token $v_i$ as tokenized by $\mathcal{M}$, and $Op(\cdot)$ denotes the operators present in any given token using CodeNet tokenizer, we allow the substitute tokens to have an extra or a missing operator (akin to typos in the natural channel of code).
\begin{equation}
    |Op(v_i)|-1 \le |Op(s_i)| \le |Op(v_i)|+1
\label{eq:constraint2}
\end{equation}
Let $C(\cdot)$ denote the code token class (identifiers, keywords, and arguments) of a token. We maintain the alignment between between $v_i$ and the potential substitute $s_i$ as follows.
\begin{equation}
    C(v_i) = C(s_i)\; \textnormal{and}\; |C(v_i)| = |C(s_i)|
\label{eq:constraint1}
\end{equation}
The above code constraints maintain the code fluency and the code consistency of $\mathcal{X}_{adv}$ and significantly reduce the search space for finding adversarial examples.

\vspace{5pt}
\noindent\textit{\textbf{Substitutions.}} We allow two types of substitutions of vulnerable tokens to generate adversarial examples: (i) \textit{Operator (character) level substitution} -- only an operator is inserted/replaced/deleted; and (ii) \textit{Token-level substitution}. We use the reduced search space $S$ and iteratively substitute, until the adversary's goal is met. We only allow replacing upto $p\%$ of the vulnerable tokens/characters to limit the number of perturbations. We also maintain the cosine similarity between the input text $\mathcal{X}$ and the adversarially perturbed text $\mathcal{X}_{adv}$ above a certain threshold (Equation \ref{eq:sim}). The complete algorithm is given in Algorithm~\ref{algo}. {\model} maintains minimal perturbation, code fluency, and code consistency between the input and the adversarial code snippet.

\begin{algorithm}[tb]
\SetAlgoLined
\SetNoFillComment
\textbf{Input:} Code $\mathcal{X}$; Victim model $F$; Maximum perturbation $\theta$; Similarity $\epsilon$; Performance Drop $\phi$\\
\textbf{Output:} Adversarial Example $\mathcal{X}_{adv}$\\
\textbf{Initialize:} $\mathcal{X}_{adv} \leftarrow \mathcal{X}$\\
\SetAlgoLined
% //~~\CommentSty{Training}\\
//~\CommentSty{Find vulnerable tokens `V'}\\
\For{$x_i$ in $\mathcal{M}(\mathcal{X})$}{
Calculate $I_{x_i}$ acc. to Eq.(\ref{eq:infl})}
$V \leftarrow \textnormal{Rank}(x_i)$ based on $I_{x_i}$\\
//~\CommentSty{Find substitutes `S'}\\
\For{$v_i$ in $V$}{
    $S \leftarrow \textnormal{Filter}(v_i)$ subject to Eqs.(\ref{eq:constraint2}), (\ref{eq:constraint1})\\
    \For{$s_j$ in $S$}{
        //~\CommentSty{Attack the victim model}\\
        $\mathcal{X}_{adv} = [x_1,...,x_{i-1},s_j,...,x_m]$ \\
        \If {$Q(F(\mathcal{X})) - Q(F(\mathcal{X}_{adv})) \ge \phi$ and $\text{Sim}(\mathcal{X}, \mathcal{X}_{adv}) \ge \epsilon$ and $||\mathcal{X}_{adv} - \mathcal{X}|| \le \theta$}{
        \Return{$\mathcal{X}_{adv}$}  //~\CommentSty{Success}
        } 
    }
    //~\CommentSty{One perturbation}\\
    $\mathcal{X}_{adv} \leftarrow
         [x_1,...x_{i-1},s_j,..x_m]$ 
}
\Return{}
% \Return{$\mathcal{X}_{adv}$} //~\CommentSty{Failed}
\caption{{\model}: Generating adversarial examples for Code}
\label{algo}
\end{algorithm}

\section*{Experiments}

We study the following research questions:
\begin{itemize}
    \item \textbf{RQ1:} How effective and transferable are the attacks generated using {\model} to different downstream tasks and programming languages?
    \item \textbf{RQ2:} How is the quality of adversarial samples generated using {\model}?
    \item \textbf{RQ3:} Is {\model} effective when we limit the number of allowed perturbations?
    \item \textbf{RQ4:} What is the impact of different components on the performance of {\model}?
\end{itemize}

\subsubsection{Downstream Tasks and Datasets}
We evaluate the transferability of {\model} across different sequence to sequence downstream tasks and in different programming languages: (i) \textit{Code Translation}\footnote{https://github.com/eclipse/jgit/,~http://lucene.apache.org/, http://poi.apache.org/, https://github.com/antlr/} involves translating between C\# and Java and vice-versa, (ii) \textit{Code Repair} automatically fixes bugs in Java functions. We use the `small' dataset \cite{tufano2019empirical}, (iii) \textit{Code Summarization} involves generating natural language summary for a given code. We use Python, Java, and PHP from the CodeSearchNet dataset \cite{husain2019codesearchnet}. (See Appendix A for details).

\subsubsection{Victim Models} We pick a representative method from different categories for our experiments: (i) \textit{CodeT5}: Pre-trained {encoder-decoder} transformer-based PL model \cite{wang2021codet5}, (ii) \textit{CodeBERT}: Bimodal pre-trained PL model \cite{feng2020codebert}, (iii) \textit{GraphCodeBERT}: Pre-trained {graph} PL model \cite{guo2020graphcodebert}, (iv) \textit{RoBERTa}: Pre-trained NL model \cite{guo2020graphcodebert}. (See Appendix A for details).

\subsubsection{Baseline Models}
Since {\model} operates in the natural channel of code, we compare with two state-of-the-art adversarial NLP baselines for a fair comparison: (i) {TextFooler}: Uses synonyms, Part-Of-Speech checking, and semantic similarity to generate adversarial text \cite{jin2020bert}, (ii) {BERT-Attack}: Uses a pre-trained BERT masked language model to generate adversarial text \cite{li2020bert}.

\subsubsection{Evaluation Metrics}\label{sec:eval}
\begin{table*}[h]
    \small
    \centering
    \begin{tabular*}{\textwidth}{ccc|cccc|ccc}
        \cline{1-10}
        \rule{0pt}{2.5ex}\textbf{Task} & \multirow{2}{1cm}{\tb{Victim Model}} & \multirow{2}{1cm}{\tb{Attack Method}}  & \multicolumn{4}{c|}{\tb{Attack Effectiveness}} & \multicolumn{3}{c}{\tb{Attack Quality}} \\[0.05cm]
        \cline{4-10}
        & & & \tb{Before} & \textbf{After} & $\Delta_{drop}$ & \textbf{Success\%} & \textbf{\#Queries} & \textbf{\#Perturb} & \textbf{CodeBLEU$_q$} \\ [0.05cm]
        \cline{1-10}
        \rule{0pt}{2.5ex}\multirow{9}{1.5cm}{Translate (Code-Code)} & \multirow{3}{*}{CodeT5} & TextFooler & \multirow{3}{*}{73.99} &  68.08 & 5.91 & 28.29 & \ul{94.95} & \ul{2.90} & \ul{63.19} \\
            & & BERT-Attack & & \ul{63.01} & \ul{10.98} & \ul{75.83} & 163.5 & 5.28 & 62.52 \\
            & & {\model}  &  & \tb{61.72} & \tb{12.27} & \tb{89.3} &  \tb{36.84} & \tb{2.55} & \tb{65.91} \\[0.05cm]
        \cline{2-10}              
            \rule{0pt}{2.5ex} & \multirow{3}{*}{CodeBERT} & TextFooler & \multirow{3}{*}{71.16} & 60.45 & 10.71 & 49.2 & \ul{73.91} & \ul{1.74} & \ul{66.61}\\
            & & BERT-Attack &  & \ul{58.80} & \ul{12.36} & \ul{70.1} & {290.1} & 5.88 & 52.14 \\
            & & {\model}  &  & \textbf{54.14} & \tb{17.03} & \tb{97.7} & \tb{26.43} & \tb{1.68} & \tb{66.89}\\[0.05cm]
        \cline{2-10}
            \rule{0pt}{2.5ex} & \multirow{3}{1cm}{\centering GraphCode-BERT} & Textfooler & \multirow{3}{*}{66.80} & 46.51 & 20.29 & 38.70 & \ul{83.17} & \ul{1.82} & \ul{63.62} \\
            & & BERT-Attack &  & \textbf{36.54} & \tb{30.26} & \ul{94.33} & {175.8} & 6.73 & 52.07\\
            & & {\model}  &   & \ul{38.81} & \ul{27.99} & \tb{98} & \tb{20.60} & \tb{1.64} & \tb{65.39}\\ [0.05cm]
        \cline{1-10}
        
    \rule{0pt}{2.5ex} \multirow{9}{1.5cm}{Repair (Code-Code)} & \multirow{3}{*}{CodeT5} & Textfooler & \multirow{3}{*}{61.13} & 57.59 & 3.53 & 58.84 & 90.50 & \ul{2.36} & \tb{69.53}\\
            &   & BERT-Attack & & \textbf{52.70} & \tb{8.43} & \ul{94.33} & \ul{262.5} & 15.1 & 53.60\\
            &   & {\model}  & & \ul{53.21} & \ul{7.92} & \tb{99.36} & \tb{30.68} & \tb{2.11} & \ul{69.03}\\ [0.05cm]
        \cline{2-10}
            \rule{0pt}{2.5ex} & \multirow{3}{*}{CodeBERT} & Textfooler & \multirow{3}{*}{61.33} & 53.55 & 7.78 & 81.61 & \ul{45.89} & \ul{2.16} & \tb{68.16}\\
            &   & BERT-Attack & & \textbf{51.95} & \tb{9.38} & \ul{95.31} & 183.3 & 15.7 & 61.95\\
            &   & {\model}  & & \ul{52.02} & \ul{9.31} & \tb{99.39} & \tb{25.98} & \tb{1.64} & \ul{68.05}\\ [0.05cm]
        \cline{2-10}
            \rule{0pt}{2.5ex} & \multirow{3}{1cm}{\centering GraphCode-BERT} & Textfooler & \multirow{3}{*}{62.16} & 54.23 & 7.92 & 78.92 & \ul{51.07} & \ul{2.20} & \tb{67.89}\\
            &   & BERT-Attack & & \ul{53.33} & \ul{8.83} & \ul{96.20} & 174.1 & 15.7 & 53.66 \\
            &   & {\model}  & & \textbf{51.97} & \tb{10.19} & \tb{99.52} & \tb{24.67} & \tb{1.67} & \ul{66.16}\\ [0.05cm]
    \cline{1-10}
    
    \rule{0pt}{2.5ex} \multirow{9}{1.5cm}{Summarize (Code-NL)} & \multirow{3}{*}{CodeT5} & TextFooler & \multirow{3}{*}{20.06} & 14.96 & 5.70 & 64.6 & \ul{410.15} & \tb{6.38} & \tb{53.91}\\
          &  & BERT-Attack & & \ul{11.96} & \ul{8.70} & \ul{78.4} & 1014.1 & \ul{7.32} & 51.34 \\
          &  & {\model}  & & \tb{11.06} & \tb{9.59} & \tb{82.8} & \tb{314.87} & 10.1 & \ul{52.67}\\ [0.05cm]
        \cline{2-10} 
        \rule{0pt}{2.5ex} & \multirow{3}{*}{CodeBERT} & Textfooler & \multirow{3}{*}{19.76} & 14.38 & 5.37 & \ul{61.10} & \ul{358.43} & \ul{2.92} & \tb{54.10}\\
        & & BERT-Attack &  & \ul{11.30} & \ul{8.35} & {56.47} & 1912.6 & 15.8 & 46.24\\
        & & {\model}  &   & \tb{10.88} & \tb{8.87} & \tb{88.32} & \tb{204.46} & \tb{2.57} & \ul{52.95}\\ [0.05cm]
        \cline{2-10}
        \rule{0pt}{2.5ex} & \multirow{3}{*}{RoBERTa} & TextFooler &  \multirow{3}{*}{19.06} & 14.06 & 4.99 & \ul{62.60} & \ul{356.68} & \ul{2.80} & \tb{54.11}\\
        & & BERT-Attack &  & \ul{11.34} & \ul{7.71} & {60.46} & 1742.3 & 17.1 & 46.95\\
        & & {\model}  &  & \tb{10.98} & \tb{8.08} & \tb{87.51} & \tb{183.22} & \tb{2.62} & \ul{53.03}\\ [0.05cm]
    \cline{1-10}
    \end{tabular*}
    \caption{Results on translation (C\#-Java), repair (Java-Java), and summarization (PHP) tasks. The performance is measured in CodeBLEU for Code-Code tasks and in BLEU for Code-NL task. The best result is in {boldface}; the next best is {underlined}.}
    \label{tab:code-code}
\end{table*}

We evaluate the \textit{effectiveness} and the \textit{quality} of the generated adversarial code.

\vspace{5pt}
\noindent\textbf{\textit{Attack Effectiveness.}} 
% To measure the effectiveness of the adversarial attacks on sequence-to-sequence tasks, we define the following metric. 
% To measure the effectiveness on sequence-to-sequence tasks, 
We define the following metric.

\begin{itemize}
    \item \textbf{$\Delta_{drop}$:} We measure the drop in the downstream performance \textit{before} and \textit{after} the attack using CodeBLEU \cite{ren2020codebleu} and BLEU \cite{papineni-etal-2002-bleu}. We define 
    \begin{equation*}
    \Delta_{drop} = Q_{\textnormal{before}}-Q_{\textnormal{after}} = Q(F(\mathcal{X}),\mathcal{Y})-Q(F(\mathcal{X}_{adv}, \mathcal{Y})    
    \end{equation*}
    where $Q = {\{\textnormal{CodeBLEU, BLEU}\}}$; $\mathcal{Y}$ is the ground truth output; $F$ is the pre-trained victim PL model, $\mathcal{X}_{adv}$ is the adversarial code sequence generated after perturbing the original input source code $\mathcal{X}$. CodeBLEU measures the quality of the \textit{generated} code snippet for code translation and code repair, and BLEU measures the quality of the \textit{generated} natural language code summary when compared to the ground truth. 
    \item \textbf{Success \%:} Computes the \% of successful attacks as measured by $\Delta_{drop}$. The higher the value, the more \textit{effective} is the adversarial attack.
\end{itemize}

\noindent\textbf{\textit{Attack Quality.}}  The following metric measures the quality of the generated adversarial code across three dimensions: (i) efficiency, (ii) imperceptibility, and (iii) code consistency.
\begin{itemize}
\item \textbf{\# Queries:} Under a black-box setting, the adversary can query the victim model to check for changes in the output logits. The lower the average number of queries required per sample, the more \textit{efficient} is the adversary. 
\item \textbf{\# Perturbation:} The number of tokens changed on an average to generate an adversarial code. The lower the value, the more \textit{imperceptible} the attack will be.
\item \tb{CodeBLEU$_q$:} Measures the consistency of the adversarial code using $\tb{\textnormal{\tb{CodeBLEU}}$_q$} = \textnormal{CodeBLEU}(\mathcal{X}, \mathcal{X}_{adv})$; where $\mathcal{X}_{adv}$ is the adversarial code sequence generated after perturbing the original input source code $\mathcal{X}$. The higher the {CodeBLEU$_q$}, the more \textit{consistent} the adversarial code is with the original source code. 
\end{itemize}

\subsubsection{Implementation Details} The model is implemented in PyTorch. We use the publicly available pre-trained CodeBERT (MLM) masked model as the adversarial code generator. We select the top 50 predictions for each vulnerable token as the initial search space and allow attacking a maximum of 40\% of code tokens. The cosine similarity threshold between the original code and adversarially generated code is set to 0.5. 
% Since {\model} requires no training, we attack the victim models on the test set.
As victim models, we use the publicly available fine-tuned checkpoints for CodeT5 and fine-tune CodeBERT, GraphCodeBERT, and RoBERTa on the related downstream tasks. We use a batch-size of 256. All experiments were conducted on a $48$ GiB RTX 8000 GPU. The source code for {\model} can be found at {https://github.com/reddy-lab-code-research/CodeAttack}.

\subsection{RQ1: Effectiveness of {\model}}\label{sec:results_rq1}
We test the effectiveness and transferability of the generated adversarial samples on three different sequence-to-sequence tasks (Code Translation, Code Repair, and Code Summarization). We generate adversarial code for four different programming languages (C\#, Java, Python, and PHP), and attack four different pre-trained PL models (CodeT5, GraphCodeBERT, CodeBERT, and Roberta). The results for C\#-Java translation task and for the PHP code summarization task are shown in Table \ref{tab:code-code}. (See Appendix A for Java-C\# translation and Python and Java code summarization tasks). {\model} has the highest success\% compared to other adversarial NLP baselines. {\model} also outperforms the adversarial baselines, BERT-Attack and TextFooler, in 6 out of 9 cases -- the average $\Delta_{drop}$ using {\model} is around $20\%$ for code translation and $10\%$ for code repair tasks, respectively. For code summarization, {\model} reduces BLEU by almost 50\% for all the victim models. As BERT-Attack replaces tokens indiscriminately, its $\Delta_{drop}$ is higher in some cases but its attack quality is the lowest.

\subsection{RQ2: Quality of Attacks Using {\model}}\label{sec:results_rq2}
\vspace{5pt}
\noindent\textbf{Quantitative Analysis.} 
\begin{figure*}
    \centering
    \includegraphics[width=0.99\textwidth]{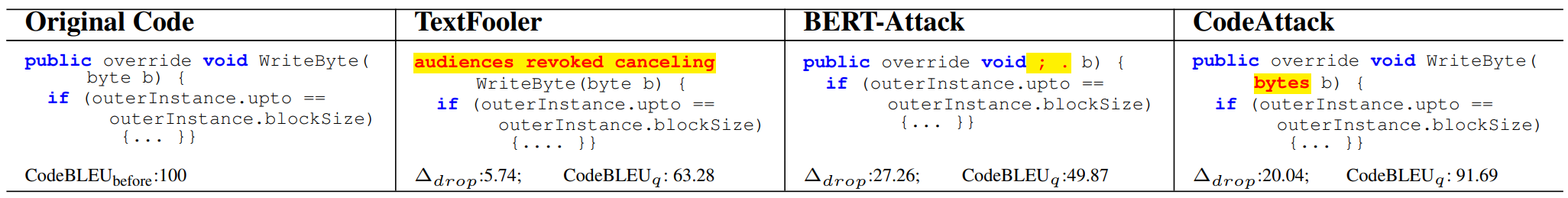}
    \caption{Qualitative examples of adversarial codes on C\#-Java Code Translation task. (See Appendix A for more examples).}
    \label{fig:qual_comp}
\end{figure*}
% We measure the quality of the adversarial attack across three dimensions - (i) efficiency, (ii) imperceptibility, and (iii) the code consistency of the generated adversarial samples using metrics defined earlier. 
Compared to the other adversarial NLP models, {\model} is the most efficient as it requires the lowest number of queries for a successful attack (Table~\ref{tab:code-code}). {\model} is also the least perceptible as the average number of perturbations required are 1-3 tokens in 8 out of 9 cases. The code consistency of adversarial samples, as measured by CodeBLEU$_q$, generated using {\model} is comparable to TextFooler which has a very low success rate. {\model} has the best overall performance.

\vspace{2pt}
\noindent\textbf{Qualitative Analysis.} 
Figure \ref{fig:qual_comp} presents qualitative examples of the generated adversarial code snippets from different attack models. Although TextFooler has a slightly better CodeBLEU$_q$ score when compared to {\model} (as seen from Table~\ref{tab:code-code}), it replaces keywords with closely related natural language words (\texttt{public} $\to$ \texttt{audiences}; \texttt{override} $\to$ \texttt{revoked}, \texttt{void} $\to$ \texttt{cancelling}). BERT-Attack has the lowest CodeBLEU$_q$ and substitutes tokens with seemingly random words. Both TextFooler and BERT-Attack have not been designed for programming languages. {\model} generates more meaningful adversarial code samples by replacing vulnerable tokens with variables and operators which are imperceptible and consistent. 

\begin{figure}
    \centering
    \includegraphics[width=0.45\textwidth]{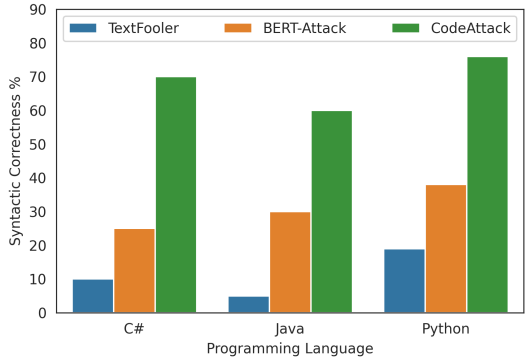}
    \caption{Syntactic correctness of adversarial code on C\#, Java, and Python demonstrating attack quality.}
    \label{fig:syntax}
\end{figure}

\vspace{2pt}
\noindent\textbf{Syntactic correctness.} Syntactic correctness of the generated adversarial code is a useful criteria for evaluating the attack quality even though {\model} and other PL models primarily operate in the natural channel of code, \textit{i.e.}, they are concerned with code understanding for humans and not with the execution or compilation of the code. The datasets described earlier consist of code snippets and cannot be compiled. Therefore, we generate adversarial code for C\#, Java, and Python using TextFooler, BERT-Attack, and {\model} and ask 3 human annotators, familiar with these languages to verify the syntax manually. We randomly sample 60 generated adversarial codes for all three programming languages for evaluating each of the above methods. {\model} has the highest average syntactic correctness for C\# (70\%), Java (60\%), and Python (76.19\%) followed by BERT-Attack and TextFooler (Figure~\ref{fig:syntax}), further highlighting the need for a code-specific adversarial attack.

\subsection{RQ3: Limiting Perturbations Using {\model}} 

\begin{figure*}[h]
\begin{subfigure}[]{0.24\linewidth}
    \includegraphics[width=\textwidth]{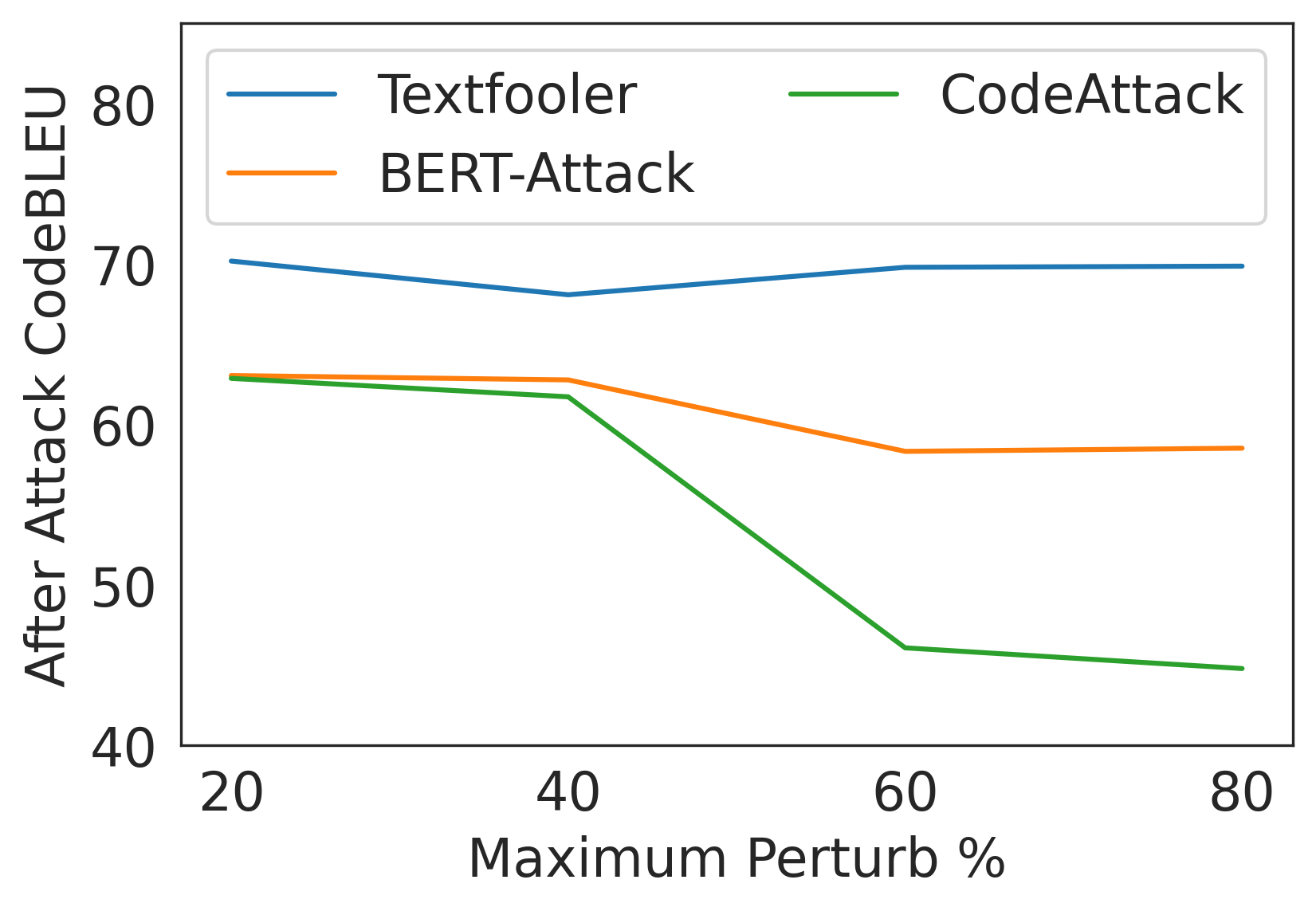}
    \caption{CodeBLEU$_\textnormal{after}$}
    \label{fig:exp_after_cb}
    \end{subfigure}
    \begin{subfigure}[]{0.24\linewidth}
    \includegraphics[width=\textwidth]{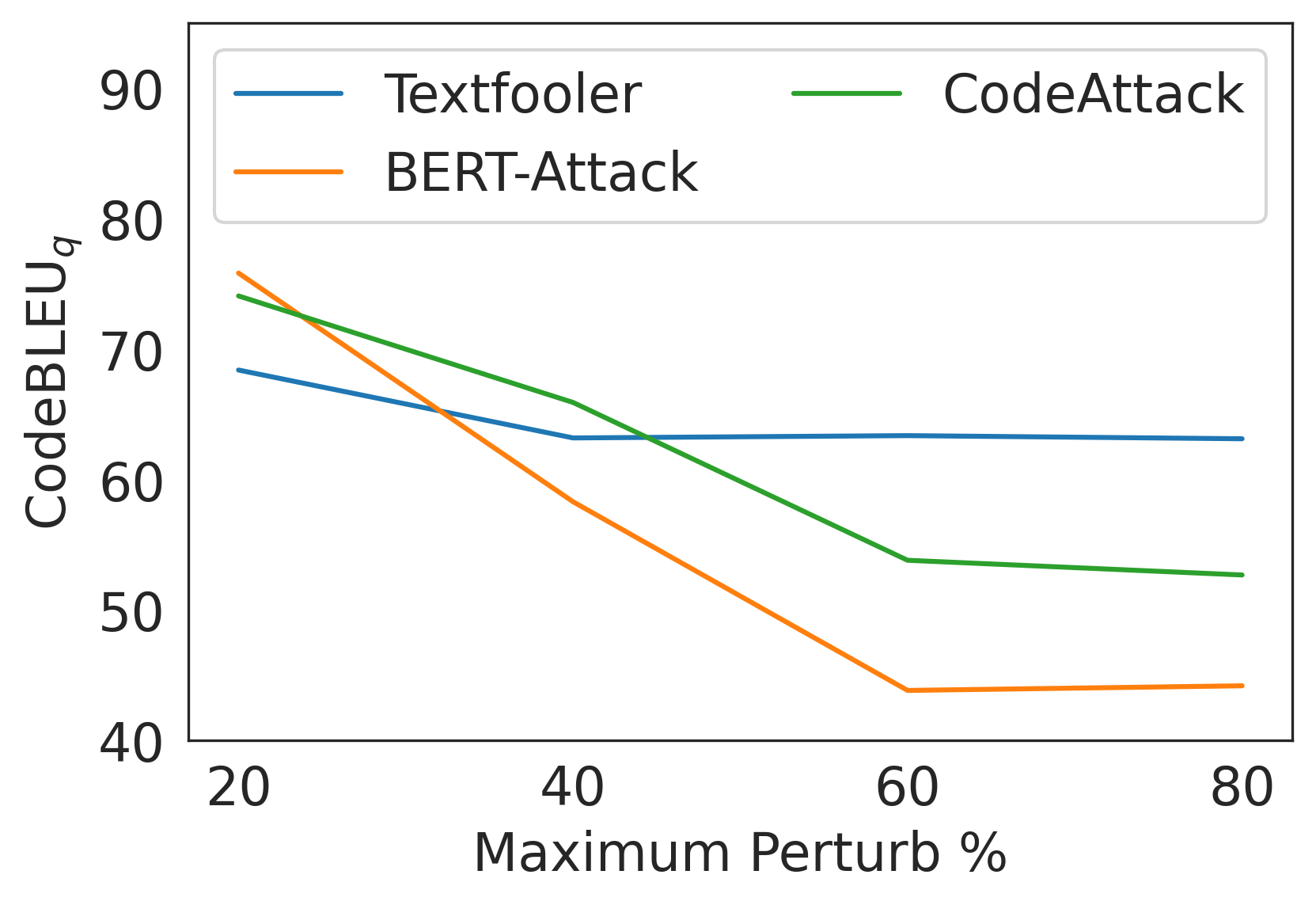}
    \caption{CodeBLEU$_q$}
    \label{fig:exp_atk_qual}
    \end{subfigure}
    \begin{subfigure}[]{0.24\linewidth}
    \includegraphics[width=\textwidth]{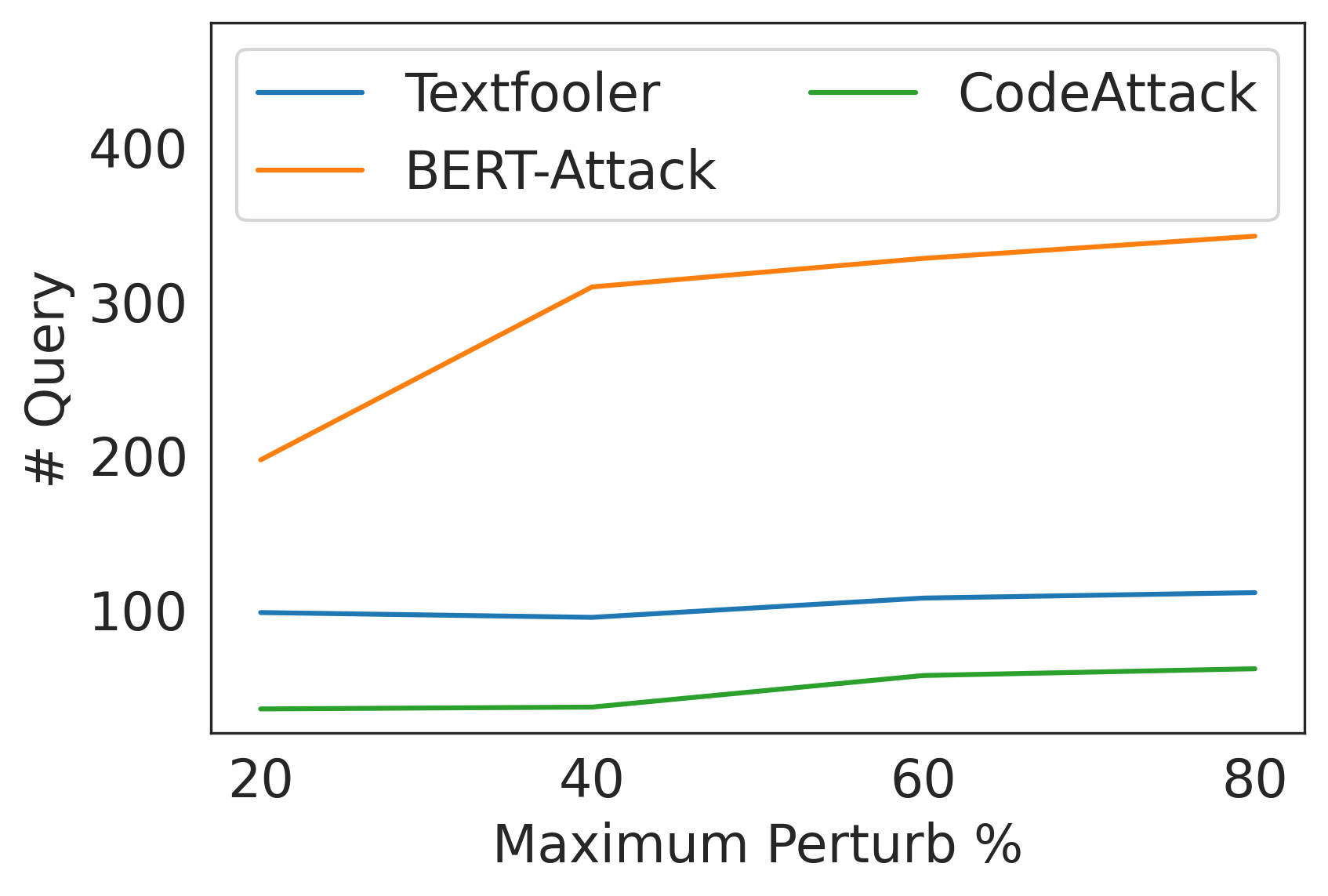}
    \caption{Average \#Queries}
    \label{fig:exp_avg_query}
    \end{subfigure}
    \begin{subfigure}[]{0.24\linewidth}
    \includegraphics[width=\textwidth]{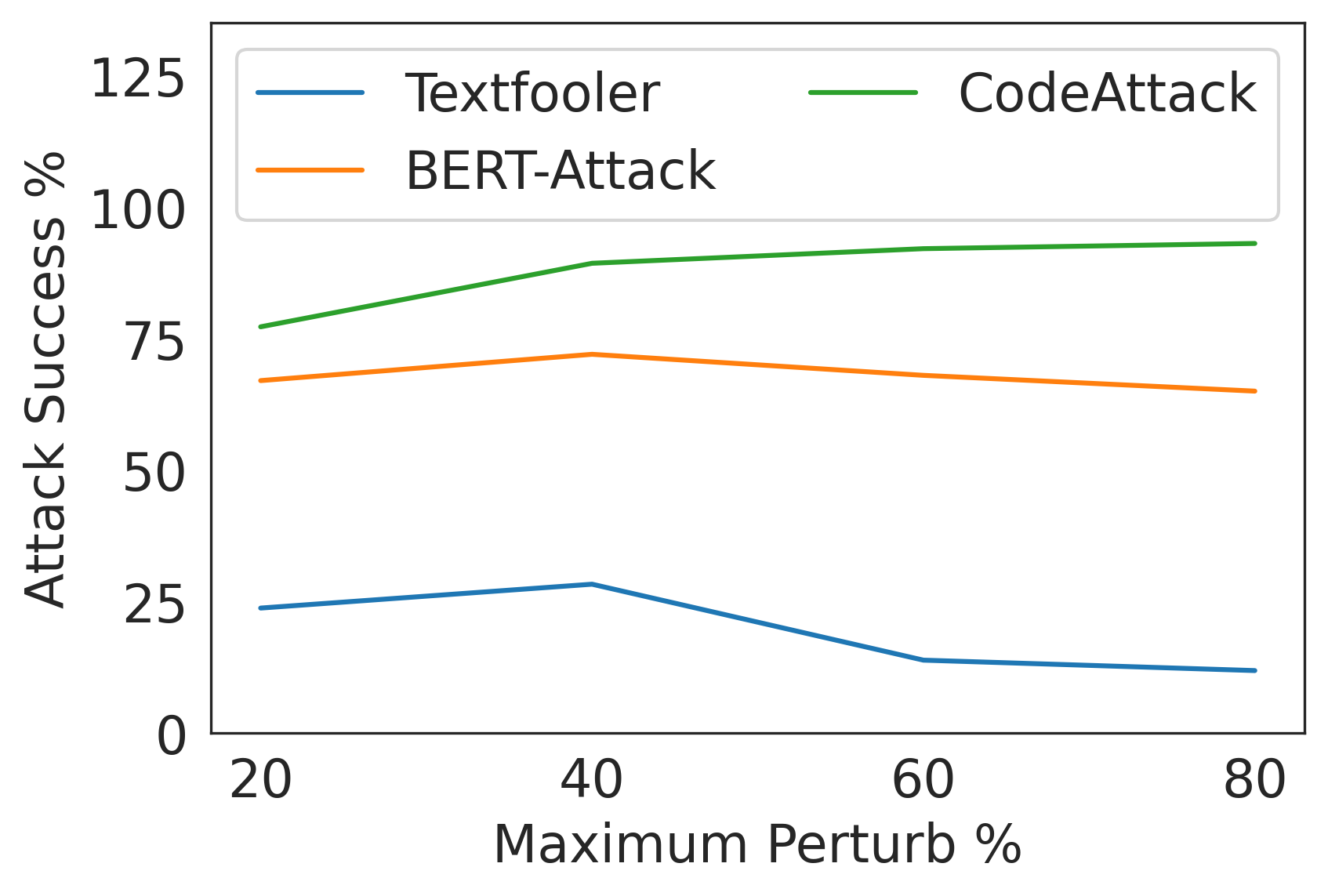}
    \caption{Attack\%}
    \label{fig:exp_success_rate}
    \end{subfigure}
    \caption{Varying the perturbation \% to study attack effectiveness on CodeT5 for the code translation task (C\#-Java).}
    \label{fig:exp_perturb}
\end{figure*}

We restrict the number of perturbations when attacking a pre-trained PL model to a strict limit, and study the effectiveness of {\model}. From Figure~\ref{fig:exp_after_cb}, we observe that as the perturbation \% increases, the CodeBLEU$_{\textnormal{after}}$ for {\model} decreases but remains constant for TextFooler and only slightly decreases for BERT-Attack. We also observe that although CodeBLEU$_q$ for {\model} is the second best (Figure~\ref{fig:exp_atk_qual}), it has the highest attack success rate (Figure~\ref{fig:exp_success_rate}) and requires the lowest number of queries for a successful attack (Figure~\ref{fig:exp_avg_query}). This shows the efficiency of {\model} and the need for code-specific adversarial attacks.

\subsection{RQ4: Ablation Study}\label{sec:ablation}

\begin{figure*}[h]
\begin{subfigure}[]{0.24\linewidth}
    \includegraphics[width=\textwidth]{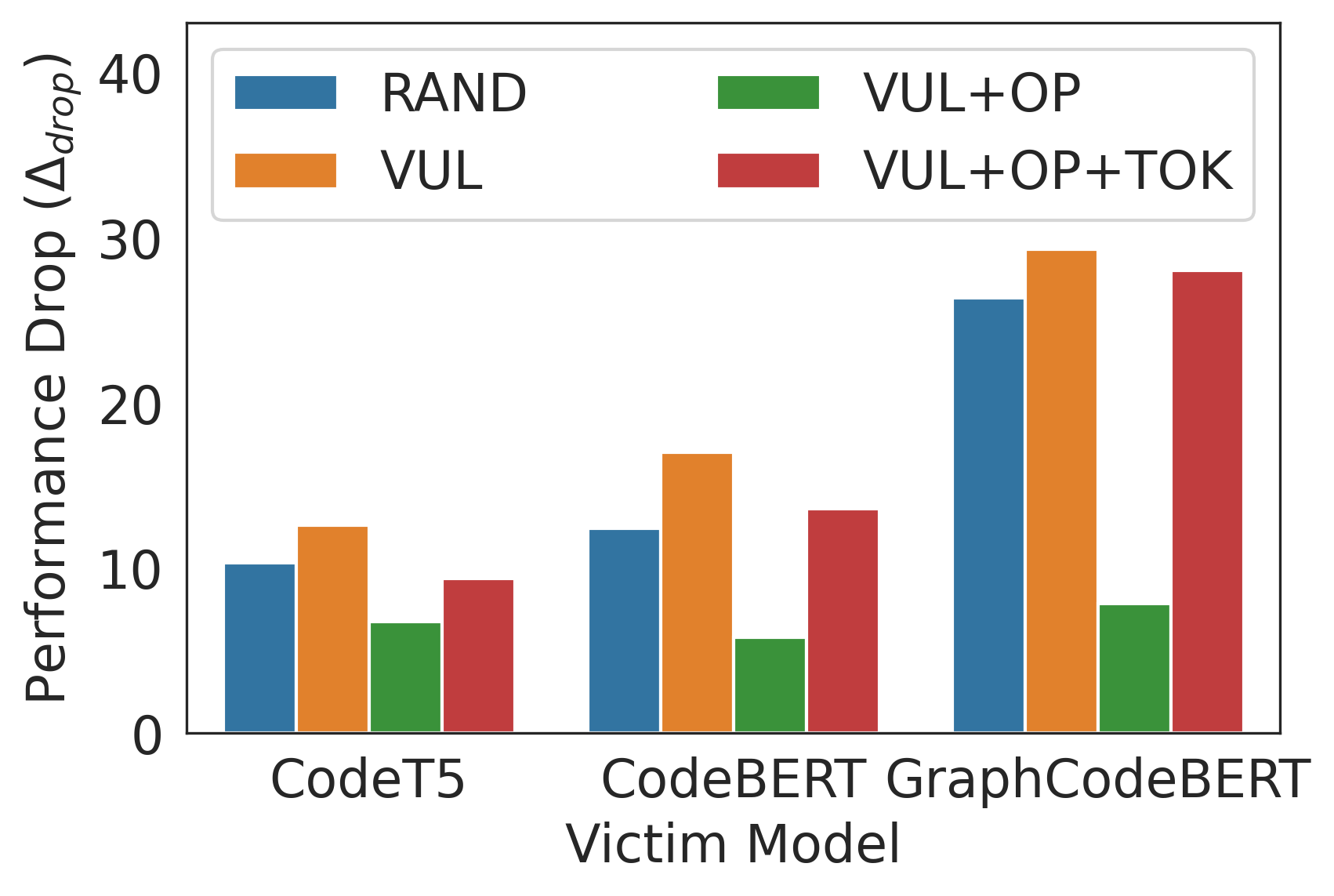}
    \caption{Performance Drop}
    \label{fig:ablation_perf_drop}
    \end{subfigure}
    \begin{subfigure}[]{0.24\linewidth}
    \includegraphics[width=\textwidth]{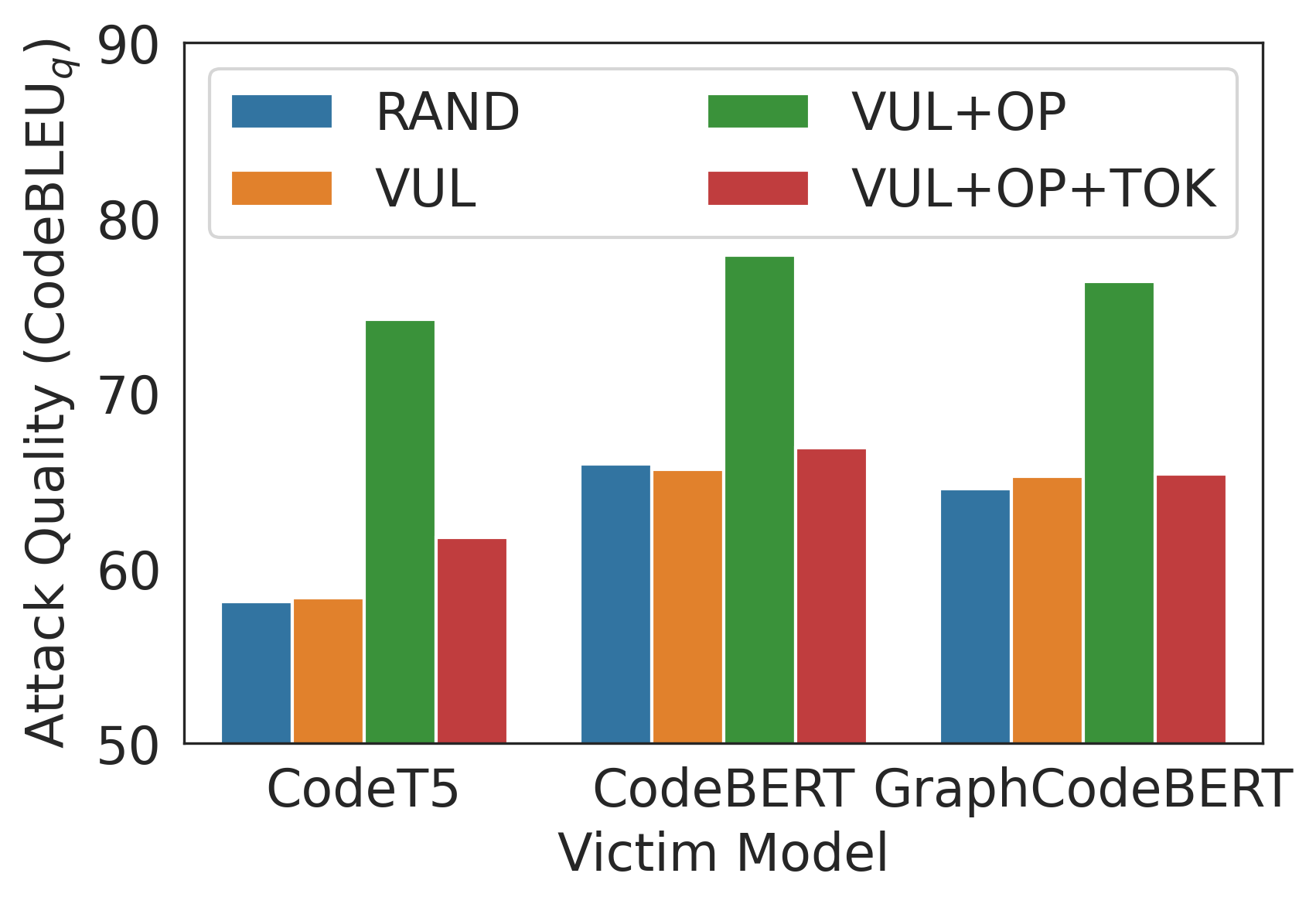}
    \caption{CodeBLEU$_q$}
    \label{fig:ablation_atk_qual}
    \end{subfigure}
    \begin{subfigure}[]{0.24\linewidth}
    \includegraphics[width=\textwidth]{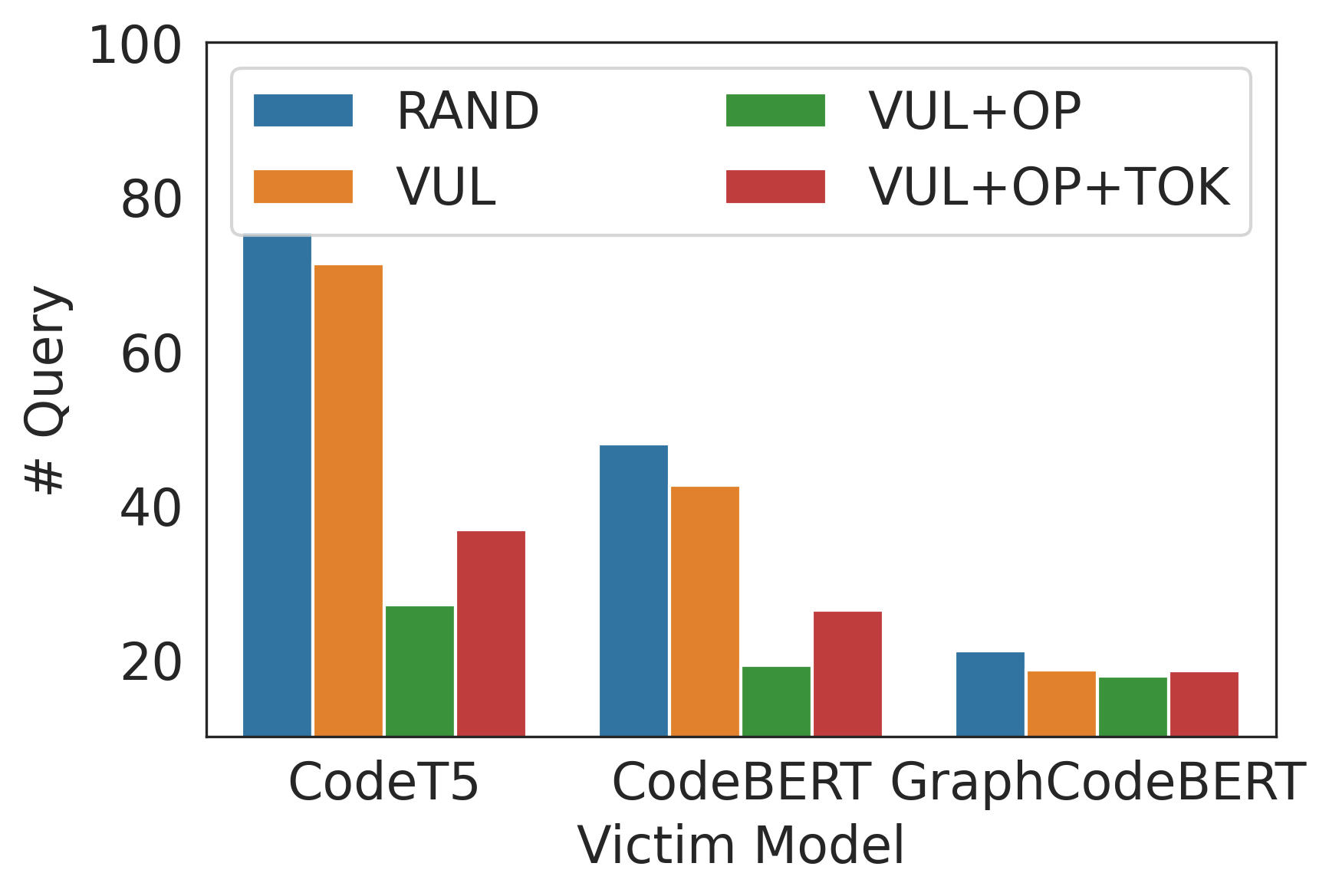}
    \caption{\# Queries}
    \label{fig:ablation_avg_query}
    \end{subfigure}
    \begin{subfigure}[]{0.24\linewidth}
    \includegraphics[width=\textwidth]{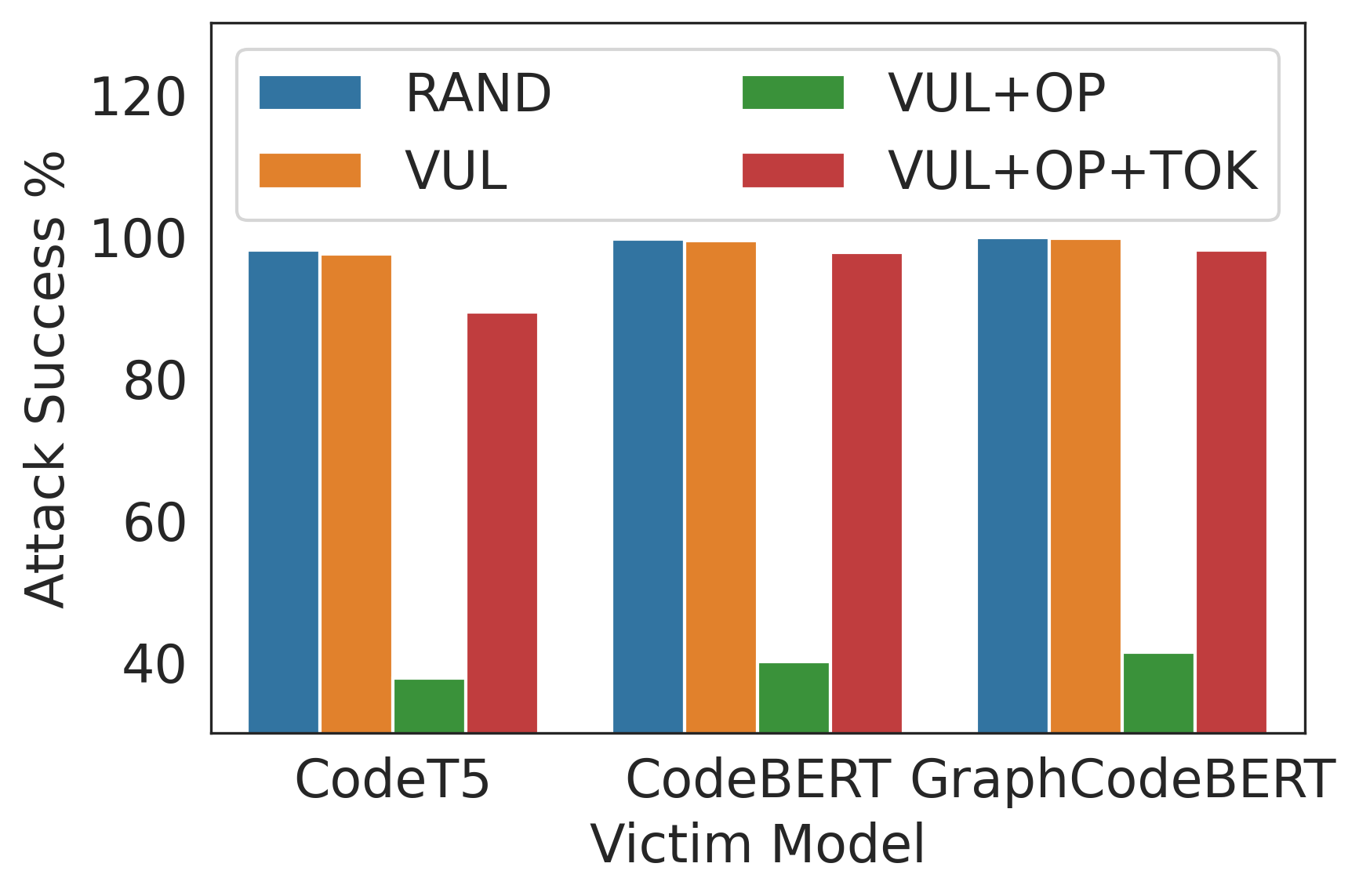}
    \caption{Average Success Rate}
    \label{fig:ablation_success_rate}
    \end{subfigure}
    \caption{Ablation Study for Code Translation (C\#-Java): Performance of different components of {\model} with random (RAND) and vulnerable tokens (VUL) and two code-specific constraints: (i) Operator level (OP), and (ii) Token level (TOK).}
    \label{fig:ablation}
\end{figure*}

% We conduct an ablation study to evaluate the importance of selecting vulnerable tokens and applying code-specific constraints to maintain the syntax of the perturbed code. 
% The results of the ablation study for the C\#-Java code translation task can be seen in Figure~\ref{fig:ablation}. 
 
\paragraph{Importance of Vulnerable Tokens.}  We create a variant, {\model}$_\texttt{RAND}$, which randomly samples tokens from the input code for substitution. We define another variant, {\model}$_\texttt{VUL}$, which finds vulnerable tokens based on logit information and attacks them, albeit without any constraints. As can be seen from Figure~\ref{fig:ablation_perf_drop}, attacking random tokens is not as effective as attacking vulnerable tokens. Using {\model}$_\texttt{VUL}$ yields greater $\Delta_{drop}$ and requires fewer number of queries when compared to {\model}$_\texttt{RAND}$, across all three models at similar CodeBLEU$_q$ (Figure~\ref{fig:ablation_atk_qual}) and success \% (Figure~\ref{fig:ablation_success_rate}). 
 
\paragraph{Importance of Code-Specific Constraints.} We find vulnerable tokens and apply two types of constraints: (i) Operator level constraint ({\model}$_\texttt{OP}$), and (ii) Token level constraint ({\model}$_\texttt{TOK}$). Only applying the operator level constraint results in lower attack success\% (Figure~\ref{fig:ablation_success_rate}) and a lower $\Delta_{drop}$ (Figure~\ref{fig:ablation_perf_drop}) but a much higher CodeBLEU$_q$. This is because we limit the changes only to operators resulting in minimal changes. On applying both operator level and token level constraints together, the $\Delta_{drop}$ and the attack success\% improve significantly. (See Appendix A for qualitative examples.)
 
Overall, the final model, {\model}, consists of {\model}$_\texttt{VUL}$, {\model}$_\texttt{OP}$, and {\model}$_\texttt{TOK}$, has the best trade-off across $\Delta_{drop}$, attack success \%, CodeBLEU$_q$, and \#Queries for all pre-trained PL victim models.

\paragraph{Human Evaluation.} 
We sample 50 original and perturbed Java and C\# code samples and shuffle them to create a mix. We ask 3 human annotators, familiar with the two programming languages, to classify the codes as either original or adversarial by evaluating the source codes in their natural channel. On average, 72.1\% of the given codes were classified as original. We also ask them to read the given adversarial codes and rate their code understanding on a scale of 1 to 5; where 1 corresponds to `Code cannot be understood at all'; and 5 corresponds to `Code is completely understandable'. The average code understanding for the adversarial codes was 4.14. Additionally, we provide the annotators with pairs of adversarial and original codes and ask them to rate the code consistency between the two using a scale between 0 to 1; where 0 corresponds to `Not at all consistent with the original code', and 1 corresponds to `Extremely consistent with the original code'. On average, the code consistency was 0.71.

\section*{Discussion}

Humans `summarize' code by reading function calls, focusing on information denoting the intention of the code (such as variable names) and skimming over structural information (such as \texttt{while} and \texttt{for} loops) \cite{rodeghero2014improving}. Pre-trained PL models operate in a similar manner and do not assign high attention weights to the grammar or the code structure \cite{zhang2022diet}. They treat software code as natural language \cite{hindle2016naturalness} and do not focus on compilation or execution of the input source code before processing them to generate an output \cite{zhang2022diet}. Through extensive experimentation, we demonstrate that this limitation of the state-of-the-art PL models can be exploited to generate adversarial examples in the natural channel of code and significantly alter their performance. 

We observe that it is easier to attack the code translation task rather than code repair or code summarization tasks. Since code repair aims to fix bugs in the given code snippet, it is more challenging to attack but not impossible. For code summarization, the BLEU score drops by almost 50\%. For all three tasks, CodeT5 is comparatively more robust whereas GraphCodeBERT is the most susceptible to attacks using {\model}. CodeT5 has been pre-trained on the task of `Masked Identifier Prediction' or deobsfuction \cite{lachaux2021dobf} where changing the identifier names does not have an impact on the code semantics. This helps the model avoid the attacks which involve changing the identifier names. GraphCodeBERT uses data flow graphs in their pre-training which relies on predicting the relationship between the identifiers. Since {\model} modifies the identifiers and perturbs the relationship between them, it proves to be extremely effective on GraphCodeBERT. This results in a more significant $\Delta_{drop}$ on GraphCodeBERT compared to other models for the code translation task. The adversarial examples from {\model}, although effective, can be avoided if the pre-trained PL models compile/execute the code before processing it. This highlights the need to incorporate explicit code structure in the pre-training stage to learn more robust program representations.

% Firstly, we use the CodeNet Tokenizer to eliminate substitute tokens from the search space and filter out words based on constraints described in Section~\ref{sec:substitutes}. However, CodeNet Tokenizer is prone to mis-tokenizing keywords as identifiers for some programming language. In such casea, {\model}  replaces these mis-identified keywords with other identifiers which makes the attacks more perceptible to humans. The second limitation of {\model} is that the generated adversarial example may not compile. Compiling an automatically generated code is a non-trivial task and none of the state-of-the-art models like CodeT5, CodeBERT, GraphCodeBERT, etc., address this limitation. There is potential for future work in this area.

\section*{Conclusion}
We introduce, {\model}, a black-box adversarial attack model to detect vulnerabilities of the state-of-the-art programming language models. It finds the most vulnerable tokens in a given code snippet and uses a greedy search mechanism to identify contextualized substitutes subject to code-specific constraints. Our model generates adversarial examples in the natural channel of code. We perform an extensive empirical and human evaluation to demonstrate the transferability of {\model} on several code-code and code-NL tasks across different programming languages. {\model} outperforms the existing state-of-the-art adversarial NLP models, in terms of its attack effectiveness, attack quality, and syntactic correctness. The adversarial samples generated using {\model} are efficient, effective, imperceptible, fluent, and code consistent. {\model} highlights the need for code-specific adversarial attacks for pre-trained PL models in the natural channel.

\newpage
\bibliography{aaai22}

\newpage
\appendix
\section{Appendix}
\label{appendix}

\subsection{Downstream Tasks and Datasets}
We evaluate the transferability of {\model} across different sequence to sequence downstream tasks and datasets in different programming languages.
\begin{itemize}
\item \textbf{Code Translation} involves translating one programming language to the other. The publicly available code translation datasets\footnote{http://lucene.apache.org/, http://poi.apache.org/, https://github.com/eclipse/jgit/,  https://github.com/antlr/} consists of parallel functions between Java and C\#. There are a total of 11,800 paired functions, out of which 1000 are used for testing. The average sequence length for Java functions is 38.51 tokens, and the average length for C\# functions is 46.16.

\item \textbf{Code Repair} refines code by automatically fixing bugs. The publicly available code repair dataset \cite{tufano2019empirical} consists of buggy Java functions as source and their corresponding fixed functions as target. We use the \textit{small} dataset with 46,680 train, 5,835 validation, and 5,835 test samples ($\le$ 50 tokens in each function).

\item \textbf{Code Summarization} involves generating natural language summary for a given code. We use the CodeSearchNet dataset \cite{husain2019codesearchnet} which consists of code and their corresponding summaries in natural language. We show the results of our model on Python (252K/14K/15K), Java (165K/5K/11K), and PHP (241K/13K/15K). The numbers in the bracket denote the samples in train/development/test set, respectively.
\end{itemize}

\subsection{Victim Models} 
We pick a representative method from different categories as our victim models to evaluate the attack.
\begin{itemize}
\item \textbf{CodeT5} \cite{wang2021codet5}: A unified pre-trained \textit{encoder-decoder} transformer-based PL model that leverages code semantics by using an identifier-aware pre-training objective. This is the state-of-the-art on several sub-tasks in the CodeXGlue benchmark \cite{DBLP:journals/corr/abs-2102-04664}. 
\item \textbf{CodeBERT} \cite{feng2020codebert}: A bimodal pre-trained \textit{programming language model} that performs code-code and code-NL tasks.
\item \textbf{GraphCodeBert} \cite{guo2020graphcodebert}: Pre-trained \textit{graph programming language model} that \textit{leverages code structure} through data flow graphs.
\item \textbf{RoBERTa} \cite{liu2019roberta}: Pre-trained \textit{natural language model} with state-of-art results on GLUE \cite{wang2018glue}, RACE \cite{lai2017race}, and SQuAD \cite{rajpurkar-etal-2016-squad} datasets.
\end{itemize}
We use the publicly available fine-tuned checkpoints for CodeT5 and fine-tune CodeBERT, GraphCodeBERT, and RoBERTa on the related downstream tasks.

\subsubsection{Baseline Models}
Since {\model} operates in the natural channel of code, we compare with two state-of-the-art adversarial NLP baselines for a fair comparison:
\begin{itemize}
    \item \textbf{TextFooler \cite{jin2020bert}}: Uses a combination of synonyms, Part-Of-Speech (POS) checking, and semantic similarity to generate adversarial text.
    \item \textbf{BERT-Attack \cite{li2020bert}}: Uses a  pre-trained BERT masked language model to generate adversarial examples satisfying a certain similarity threshold. 
\end{itemize}

\subsection{Results}

\paragraph{Downstream Performance and Attack Quality} We measure the CodeBLEU and $\Delta_{CodeBLEU}$ to evaluate the downstream performance for code-code tasks (code repair and code translation). The programming languages used are C\#-Java and Java-C\# for translation tasks; and Java for code repair tasks (Table~\ref{tab:appendix_results} and Table~\ref{tab:appendix_qual}). We show the results for code-NL task for code summarization in BLEU and $\Delta_{BLEU}$. We show the results for three programming languages: Python, Java, and PHP (Table~\ref{tab:appendix_results} and  Table~\ref{tab:appendix_summary}). We measure the quality of the attacks using the metric defined in \ref{sec:eval}. The results follow a similar pattern as that seen in Sections \ref{sec:results_rq1}.

\paragraph{Ablation Study: Qualitative Analysis} Table~\ref{tab:app_qual} shows the adversarial examples generated using the variants described in Section~\ref{sec:ablation}.

\begin{table*}[h]
    \small
    \centering
    \begin{tabular*}{\textwidth}{ccc|cccc|ccc}
        \cline{1-10}
        \rule{0pt}{2.5ex}\textbf{Task} & \multirow{2}{1cm}{\tb{Victim Model}} & \multirow{2}{1cm}{\tb{Attack Method}}  & \multicolumn{4}{c|}{\tb{Attack Effectiveness}} & \multicolumn{3}{c}{\tb{Attack Quality}} \\[0.05cm]
        \cline{4-10}
        & & & \tb{Before} & \textbf{After} & $\Delta_{drop}$ & \textbf{Success\%} & \textbf{\#Queries} & \textbf{\#Perturb} & \textbf{CodeBLEU$_q$} \\ [0.05cm]
        \cline{1-10}
    \rule{0pt}{2.5ex}\multirow{9}{1.5cm}{Translate (C\#-Java)} & \multirow{3}{*}{CodeT5} & TextFooler & \multirow{3}{*}{73.99} &  68.08 & 5.91 & 28.29 & \ul{94.95} & \ul{2.90} & \ul{63.19} \\
            & & BERT-Attack & & \ul{63.01} & \ul{10.98} & \ul{75.83} & 163.5 & 5.28 & 62.52 \\
            & & {\model}  &  & \tb{61.72} & \tb{12.27} & \tb{89.3} &  \tb{36.84} & \tb{2.55} & \tb{65.91} \\[0.05cm]
        \cline{2-10}              
            \rule{0pt}{2.5ex} & \multirow{3}{*}{CodeBERT} & TextFooler & \multirow{3}{*}{71.16} & 60.45 & 10.71 & 49.2 & \ul{73.91} & \ul{1.74} & \ul{66.61}\\
            & & BERT-Attack &  & \ul{58.80} & \ul{12.36} & \ul{70.1} & {290.1} & 5.88 & 52.14 \\
            & & {\model}  &  & \textbf{54.14} & \tb{17.03} & \tb{97.7} & \tb{26.43} & \tb{1.68} & \tb{66.89}\\[0.05cm]
        \cline{2-10}
            \rule{0pt}{2.5ex} & \multirow{3}{1cm}{\centering GraphCode-BERT} & Textfooler & \multirow{3}{*}{66.80} & 46.51 & 20.29 & 38.70 & \ul{83.17} & \ul{1.82} & \ul{63.62} \\
            & & BERT-Attack &  & \textbf{36.54} & \tb{30.26} & \ul{94.33} & {175.8} & 6.73 & 52.07\\
            & & {\model}  &   & \ul{38.81} & \ul{27.99} & \tb{98} & \tb{20.60} & \tb{1.64} & \tb{65.39}\\ [0.05cm]
        \cline{1-10}

    \rule{0pt}{2.5ex}\multirow{9}{1.5cm}{Translate (Java-C\#)} & \multirow{3}{*}{CodeT5} & TextFooler & \multirow{3}{*}{87.03} & 79.83 &  7. 20 & 32.3 & 62.91 & \ul{2.19}  & \tb{81.28} \\
            & & BERT-Attack & & \ul{68.81} & \ul{18.22} & \ul{86.3} & 89.99 & 2.79 & 74.52 \\
            & & {\model}  & & \tb{66.97} & \tb{20.06} & \tb{94.8} & \tb{19.85} & \tb{2.03} & \ul{75.21} \\[0.05cm]
        \cline{2-10}              
            \rule{0pt}{2.5ex} & \multirow{3}{*}{CodeBERT} & TextFooler & \multirow{3}{*}{83.48} &  73.52 & 9.96 & 55.9 & 38.57 & \ul{2.14} & \ul{73.93}\\
            & & BERT-Attack  & & \ul{67.94}  & \ul{15.5} & \ul{70.3}  & 159.1 & 5.76 & 46.82 \\
            & & {\model}  & & \tb{66.98} & \tb{16.5} & \tb{91.1} & \tb{24.42} & \tb{1.66} & \tb{76.77} \\[0.05cm]
        \cline{2-10}
            \rule{0pt}{2.5ex} & \multirow{3}{1cm}{\centering GraphCode-BERT} & Textfooler & \multirow{3}{*}{82.40} & 74.32 & 8.2 & 51.2 & 39.33 & \ul{2.03} & \ul{72.45} \\
            & & BERT-Attack & & \ul{64.87} & \ul{17.5} & \ul{76.6} & 134.3 & 5.90 & 47.47\\
            & & {\model} & & \tb{58.88} & \tb{23.5} & \tb{90.8} & \tb{23.22} & \tb{1.64} & \tb{77.33} \\ [0.05cm]
        \cline{1-10}
        
    \rule{0pt}{2.5ex} \multirow{9}{1.5cm}{Repair (small)} & \multirow{3}{*}{CodeT5} & Textfooler & \multirow{3}{*}{61.13} & 57.59 & 3.53 & 58.84 & 90.50 & \ul{2.36} & \tb{69.53}\\
            &   & BERT-Attack & & \textbf{52.70} & \tb{8.43} & \ul{94.33} & \ul{262.5} & 15.1 & 53.60\\
            &   & {\model}  & & \ul{53.21} & \ul{7.92} & \tb{99.36} & \tb{30.68} & \tb{2.11} & \ul{69.03}\\ [0.05cm]
        \cline{2-10}
            \rule{0pt}{2.5ex} & \multirow{3}{*}{CodeBERT} & Textfooler & \multirow{3}{*}{61.33} & 53.55 & 7.78 & 81.61 & \ul{45.89} & \ul{2.16} & \tb{68.16}\\
            &   & BERT-Attack & & \textbf{51.95} & \tb{9.38} & \ul{95.31} & 183.3 & 15.7 & 61.95\\
            &   & {\model}  & & \ul{52.02} & \ul{9.31} & \tb{99.39} & \tb{25.98} & \tb{1.64} & \ul{68.05}\\ [0.05cm]
        \cline{2-10}
            \rule{0pt}{2.5ex} & \multirow{3}{1cm}{\centering GraphCode-BERT} & Textfooler & \multirow{3}{*}{62.16} & 54.23 & 7.92 & 78.92 & \ul{51.07} & \ul{2.20} & \tb{67.89}\\
            &   & BERT-Attack & & \ul{53.33} & \ul{8.83} & \ul{96.20} & 174.1 & 15.7 & 53.66 \\
            &   & {\model}  & & \textbf{51.97} & \tb{10.19} & \tb{99.52} & \tb{24.67} & \tb{1.67} & \ul{66.16}\\ [0.05cm]
    \cline{1-10}
    
    \rule{0pt}{2.5ex} \multirow{9}{1.5cm}{Summarize (PHP)} & \multirow{3}{*}{CodeT5} & TextFooler & \multirow{3}{*}{20.06} & 14.96 & 5.70 & 64.6 & \ul{410.15} & \tb{6.38} & \tb{53.91}\\
          &  & BERT-Attack & & \ul{11.96} & \ul{8.70} & \ul{78.4} & 1014.1 & \ul{7.32} & 51.34 \\
          &  & {\model}  & & \tb{11.06} & \tb{9.59} & \tb{82.8} & \tb{314.87} & 10.1 & \ul{52.67}\\ [0.05cm]
        \cline{2-10} 
        \rule{0pt}{2.5ex} & \multirow{3}{*}{CodeBERT} & Textfooler & \multirow{3}{*}{19.76} & 14.38 & 5.37 & \ul{61.10} & \ul{358.43} & \ul{2.92} & \tb{54.10}\\
        & & BERT-Attack &  & \ul{11.30} & \ul{8.35} & {56.47} & 1912.6 & 15.8 & 46.24\\
        & & {\model}  &   & \tb{10.88} & \tb{8.87} & \tb{88.32} & \tb{204.46} & \tb{2.57} & \ul{52.95}\\ [0.05cm]
        \cline{2-10}
        \rule{0pt}{2.5ex} & \multirow{3}{*}{RoBERTa} & TextFooler &  \multirow{3}{*}{19.06} & 14.06 & 4.99 & \ul{62.60} & \ul{356.68} & \ul{2.80} & \tb{54.11}\\
        & & BERT-Attack &  & \ul{11.34} & \ul{7.71} & {60.46} & 1742.3 & 17.1 & 46.95\\
        & & {\model}  &  & \tb{10.98} & \tb{8.08} & \tb{87.51} & \tb{183.22} & \tb{2.62} & \ul{53.03}\\ [0.05cm]
    \cline{1-10}
    
    \rule{0pt}{2.5ex} \multirow{9}{1.5cm}{Summarize (Python)} & \multirow{3}{*}{CodeT5} & TextFooler & \multirow{3}{*}{20.36} & 12.11  &  8.25 & 90.47 & \ul{400.06} & \ul{5.26} & \tb{77.59} \\
            & & BERT-Attack & & \ul{8.22} & \ul{12.14} & \ul{97.81} & 475.61 & 6.91 & 66.27 \\
            & & {\model}  &  & \tb{7.97} & \tb{12.39} & \tb{98.50} & \tb{174.05} & \tb{5.10} & \ul{69.17} \\[0.05cm]
        \cline{2-10}              
            \rule{0pt}{2.5ex} & \multirow{3}{*}{CodeBERT} & TextFooler & \multirow{3}{*}{26.17} & \ul{22.76} & \ul{3.41} & 68.50 & 966.19 & 3.83 & \tb{75.15} \\
            & & BERT-Attack  & & 22.88 & 3.29 & \ul{84.41} & \ul{941.94} & \ul{3.35} & 56.31 \\
            & & {\model}  & & \tb{18.69} & \tb{7.48} & \tb{86.63} & \tb{560.68} & \tb{3.23} & \ul{59.11} \\[0.05cm]
        \cline{2-10}
            \rule{0pt}{2.5ex} & \multirow{3}{*}{RoBERTa} & Textfooler & \multirow{3}{*}{17.01} & 10.72 & 6.29 & 63.34 & 788.25 & 3.57 & 70.48 \\
            & & BERT-Attack & & \ul{10.66} & \ul{6.35} & \ul{74.64} & 1358.8 & 4.07 & 51.74 \\
            & & {\model} & & \tb{9.50} & \tb{7.51} & \tb{76.09} & \tb{661.75} & \tb{3.46} & \ul{61.22} \\ [0.05cm]
        \cline{1-10}
        
    \rule{0pt}{2.5ex} \multirow{9}{1.5cm}{Summarize (Java)} & \multirow{3}{*}{CodeT5} & TextFooler & \multirow{3}{*}{19.77} & 14.06 & 5.71 & 67.80 & \ul{291.82} & \tb{3.76} & \tb{90.82} \\
            & & BERT-Attack & & \ul{11.94} & \ul{7.83} & \ul{77.37} & 811.97 & 17.4 & 45.71 \\
            & & {\model}  & & \tb{11.21} & \tb{8.56} & \tb{80.80} & \tb{198.11} & \ul{7.43} & \ul{90.04} \\[0.05cm]
        \cline{2-10}              
            \rule{0pt}{2.5ex} & \multirow{3}{*}{CodeBERT} & TextFooler & \multirow{3}{*}{17.65} & 16.44 & 1.21 & 42.4 & \ul{400.78} & \ul{4.07} & \tb{90.29} \\
            & & BERT-Attack  & & \ul{15.49} & \ul{2.16} & \ul{46.51} & 1531.1 & 10.9 & 37.61 \\
            & & {\model}  & & \tb{14.69} & \tb{2.96} & \tb{73.70} & \tb{340.99} & \tb{3.27} & \ul{59.37} \\[0.05cm]
        \cline{2-10}
            \rule{0pt}{2.5ex} & \multirow{3}{*}{RoBERTa} & Textfooler & \multirow{3}{*}{16.47} & 13.23 & 3.24 & \ul{44.9} & \ul{383.36} & \ul{4.02} & \tb{90.87} \\
            & & BERT-Attack & & \ul{11.89} & \ul{4.58} & 42.59 & 1582.2 & 9.21 & 37.86 \\
            & & {\model} & & \tb{11.74} & \tb{4.73} & \tb{50.14} & \tb{346.07} & \tb{3.29} & \ul{48.48}\\ [0.05cm]
        \cline{1-10}
        
    \end{tabular*}
    \caption{Results on code translation, code repair, and code summarization tasks. The performance is measured in CodeBLEU for Code-Code tasks and in BLEU for Code-NL (summarization) task. The best result is in \tb{boldface}; the next best is \ul{underlined}.}
    \label{tab:appendix_results}
\end{table*}

\begin{table*}[h]
    \small
    \centering
    \begin{tabular*}{\textwidth}{p{4cm}|p{4cm}|p{4cm}|p{4cm}}
    \cline{1-4}
    \textbf{Original Code} & \tb{TextFooler} & \tb{BERT-Attack} & \tb{\model} \\
    \cline{1-4}
\begin{C}
public string GetFullMessage() {
  ...
  if (msgB < 0){return string.Empty;}
  ...
  return RawParseUtils.Decode(enc, raw, msgB, raw.Length);}
\end{C}
&
\begin{C}
<@\textcolor{red}{\tb{\hl{citizenship}}}@> string GetFullMessage() {
  ...
  if (msgB < 0){return string.Empty;}
  ...
  return RawParseUtils.Decode(enc, raw, msgB, raw.Length);}
\end{C} &
\begin{C}[basicstyle=\tiny]
<@\textcolor{red}{\tb{\hl{loop}}}@> string GetFullMessage() {
  ...
  if (msgB < 0){return string.Empty;}
  ...
  return <@\textcolor{red}{\tb{\hl{here q. dir, (x)}}}@> raw, msgB, raw.Length);}
\end{C} &
\begin{C}[style=CStyle,basicstyle=\tiny]
public string GetFullMessage() {
  ...
  if (msgB <@\textcolor{red}{\tb{\hl{=}}}@> 0){return string.Empty;}
  ...
  return RawParseUtils.Decode(enc, raw, msgB, raw.Length);}
\end{C}\\
\tiny{CodeBLEU$_\textnormal{before}$: 77.09} & \tiny{$\Delta_{drop}$: 18.84;\qquad CodeBLEU$_q$: 95.11} & \tiny{$\Delta_{drop}$: 15.09;\qquad CodeBLEU$_q$: 57.46} & \tiny{$\Delta_{drop}$: 21.04;\qquad CodeBLEU$_q$: 88.65}\\
\cline{1-4}
\begin{C}[style=CStyle,basicstyle=\tiny]
public override void WriteByte(byte b) {
  if (outerInstance.upto == outerInstance.blockSize) {... }}
\end{C}&
\begin{C}[style=CStyle,basicstyle=\tiny]
<@\textcolor{red}{\tb{\hl{audiences revoked canceling}}}@> WriteByte(byte b) {
  if (outerInstance.upto == outerInstance.blockSize) {.... }}
\end{C}
& 
\begin{C}[style=CStyle,basicstyle=\tiny]
public override void<@\textcolor{red}{\tb{\hl{ ; .}}}@> b) {
  if (outerInstance.upto == outerInstance.blockSize) {... }}
\end{C}&
\begin{C}[style=CStyle,basicstyle=\tiny]
public override void WriteByte(<@\textcolor{red}{\tb{\hl{bytes}}}@> b) {
  if (outerInstance.upto == outerInstance.blockSize) {... }}
\end{C}\\
\tiny{CodeBLEU$_\textnormal{before}$:100} & \tiny{$\Delta_{drop}$:5.74; \qquad CodeBLEU$_q$: 63.28} & \tiny{$\Delta_{drop}$:27.26; \qquad CodeBLEU$_q$:49.87} & \tiny{$\Delta_{drop}$:20.04;\qquad CodeBLEU$_q$: 91.69}\\
\cline{1-4}
    \end{tabular*}
    \caption{Qualitative examples of perturbed codes using TextFooler, BERT-Attack, and {\model} on Code Translation task.}
    \label{tab:appendix_qual}
\end{table*}

\begin{table*}[h]
    \small
    \centering
    \begin{tabular*}{\textwidth}{p{4cm}|p{4cm}|p{4cm}|p{4cm}}
    \cline{1-4}
    \tb{Original} & \tb{TextFooler} & \tb{BERT-Attack} & \tb{\model}\\
    \cline{1-4}

\begin{C}
protected final void fastPathOrderedEmit(U value, boolean delayError, Disposable disposable) {
    final Observer<? super V> observer = downstream;
    final ..... {
        if (q.isEmpty()) {
            accept(observer, value);
            if (leave(-1) == 0) {
                return;
            }
        } else {
            q.offer(value);
        }
    } else {
        q.offer(value);
        if (!enter()) {
            return;
        }
    }
    QueueDrainHelper.drainLoop(q, observer, delayError, disposable, this);
}

\end{C}
& 
\begin{C}
protected final <@\textcolor{red}{\tb{\hl{invalidate}}}@> fastPathOrderedEmit(U value, boolean delayError, Disposable disposable) {
  <@\textcolor{red}{\tb{\hl{finalizing}}}@> Observer < ? super V > observer = downstream;
  final .... {
    if (q.isEmpty()) {
      accept(observer, value);
      if (leave(-1) == 0) {
        return;
      }
    }
    <@\textcolor{red}{\tb{\hl{yet}}}@> {
      q.offer(value);
    }
  }
  <@\textcolor{red}{\tb{\hl{annan}}}@> {
    q.offer(value);
    than(!enter()) {
      return;
    }
  }
  QueueDrainHelper.drainLoop(q, observer, delayError, disposable, this);
}
\end{C} 
& 
\begin{C}
<@\textcolor{red}{\tb{\hl{(; period}}}@> fastPathOrderedEmit(U value, <@\textcolor{red}{\tb{\hl{this)
  0 c | / , ) ( fore pas}}}@> Observer<? super V > observer = downstream;
  final .... {
    if (q.isEmpty()) {
        accept(<@\textcolor{red}{\tb{\hl{) , 
        point 0 
        while ( leave ... ) ] a ) ] ] 
            returns 
      public . next , manager q . offer , value ) 
      \u3009 therefore 
      return draw q . offer , value ) ... ) 
      ? enter ) public }}}@>
    }
  QueueDrainHelper.drainLoop(q, observer, <@\textcolor{red}{\tb{\hl{) ) 1 0c ) ) , these ?}}}@>
 }
\end{C} 

& 
\begin{C}
protected final <@\textcolor{red}{\tb{\hl{static}}}@> fastPathOrderedEmit(<@\textcolor{red}{\tb{\hl{M}}}@> value, boolean <@\textcolor{red}{\tb{\hl{ZdelayExc}}}@>, Disposable <@\textcolor{red}{\tb{\hl{dis\_zyk}}}@>) {
  final Observer <? super V > observer = downstream;
  final .... {
    if (q.isEmpty()) {
      accept(observer, value);
      if (leave(-1) == 0) {
        <@\textcolor{red}{\tb{\hl{continue;}}}@>
      }
    } <@\textcolor{red}{\tb{\hl{catch \@}}}@> {
      q.offer(value);
    }
  } else {
        q.offer(value);
        if (!<@\textcolor{red}{\tb{\hl{exit()}}}@>) {
            return;
        }
  }
  QueueDrainHelper.drainLoop(q, observer, <@\textcolor{red}{\tb{\hl{delayInfo}}}@>, disposable, this);
}
\end{C}

\\
\cline{1-4}
Makes sure the fast-path emits in order
& This method is used to avoid the need for the fast path to emit a value to the downstream.
& period $0 | | | | | |$
& dddddsssss \\
\cline{1-4}
% \tiny{CodeBLEU$_\textnormal{before}$:100} & \tiny{$\Delta_{drop}$:5.74; \qquad CodeBLEU$_q$: 63.28} & \tiny{$\Delta_{drop}$:27.26; \qquad CodeBLEU$_q$:49.87} & \tiny{$\Delta_{drop}$:20.04;\qquad CodeBLEU$_q$: 91.69}\\
% \cline{1-4}
    \end{tabular*}
    \caption{Qualitative examples of adversarial codes and the generated summary using TextFooler, BERT-Attack, and {\model} on Code Summarization task.}
    \label{tab:appendix_summary}
\end{table*}

\begin{table*}[h]
    \small
    \centering
    \begin{tabular*}{\textwidth}{p{4cm}|p{4cm}|p{4cm}|p{4cm}}
    \cline{1-4}
    \textbf{Original Code} & \tb{{\model}$_\texttt{VUL}$} & \tb{{\model}$_{\texttt{VUL}+\texttt{OP}}$} & \tb{{\model}$_{{\texttt{VUL}+\texttt{OP}}+\texttt{TOK}}$} \\
    \cline{1-4}
\begin{C}[style=CStyle,basicstyle=\tiny]
public void AddMultipleBlanks(MulBlankRecord mbr) {
  for (int j = 0; j < mbr.NumColumns; j++) {
    BlankRecord br = new BlankRecord();
    br.Column = j + mbr.FirstColumn;
    br.Row = mbr.Row;
    br.XFIndex = (mbr.GetXFAt(j));
    InsertCell(br);
  }
}
\end{C}
&
\begin{C}[style=CStyle,basicstyle=\tiny]
<@\textcolor{red}{\tb{\hl{((}}}@>void AddMultipleBlanks(MulBlankRecord mbr) {
      for (int j <@\textcolor{red}{\tb{\hl{?}}}@> 0; j < mbr.NumColumns; j++) {
        BlankRecord br = new BlankRecord();
        br.Column = j + mbr.FirstColumn;
        br.Row = mbr.Row;
        br.XFIndex = (mbr.GetXFAt(j));
        InsertCell(br);
      }
    } 
\end{C}
&
\begin{C}[basicstyle=\tiny]
public void AddMultipleBlanks(MulBlankRecord mbr) {
  for (int j <@\textcolor{red}{\tb{\hl{>}}}@> 0; j < mbr.NumColumns; j++) {
    BlankRecord br = <@\textcolor{red}{\tb{\hl{-}}}@>new BlankRecord();
    br.Column = j + mbr.FirstColumn;
    br.Row = mbr.Row;
    br.XFIndex <@\textcolor{red}{\tb{\hl{>}}}@> (mbr.GetXFAt(j));
    InsertCell(br);
  }
}
\end{C} &
\begin{C}[style=CStyle,basicstyle=\tiny]
<@\textcolor{red}{\tb{\hl{static}}}@> void AddMultipleBlanks(MulBlankRecord mbr) {
  for (int j > 0; <@\textcolor{red}{\tb{\hl{jj}}}@> < mbr.NumColumns; j++) {
    BlankRecord br = new BlankRecord();
    br.Column = j + mbr.FirstColumn;
    br.Row = mbr.Row;
    br.XFIndex = (mbr.GetXFAt(j));
    InsertCell(br);
  }
}
\end{C}\\
\tiny{CodeBLEU$_\textnormal{before}$: 76.3} & \tiny{$\Delta_{drop}$: 7.21;\qquad CodeBLEU$_q$: 43.85} & \tiny{$\Delta_{drop}$: 5.85;\qquad CodeBLEU$_q$: 69.61} & \tiny{$\Delta_{drop}$: 12.96;\qquad CodeBLEU$_q$: 59.29}\\
\cline{1-4}
\begin{C}[style=CStyle,basicstyle=\tiny]
public string GetFullMessage() {
  byte[] raw = buffer;
  int msgB = RawParseUtils.TagMessage(raw, 0);
  if (msgB < 0) {
    return string.Empty;
  }
  Encoding enc = RawParseUtils.ParseEncoding(raw);
  return RawParseUtils.Decode(enc, raw, msgB, raw.Length);
}
\end{C}
& 
\begin{C}[style=CStyle,basicstyle=\tiny]
<@\textcolor{red}{\tb{\hl{\u0120public}}}@> string GetFullMessage() {
  byte[] raw = buffer;
  int msgB = RawParseUtils.TagMessage(raw, 0);
  if (msgB < 0) {
    return string.Empty;
  }
  Encoding enc = RawParseUtils.ParseEncoding(raw);
  return RawParseUtils.Decode(enc, <@\textcolor{red}{\tb{\hl{RAW..}}}@>, msgB, raw.Length);
}
\end{C}
&
\begin{C}[style=CStyle,basicstyle=\tiny]
public string GetFullMessage() {
  byte[] raw = buffer;
  int msgB = RawParseUtils.TagMessage(raw, 0);
  if (msgB <@\tb{\textcolor{red}{\hl{=}}}@> 0) {
    return string.Empty;
  }
  Encoding enc = RawParseUtils.ParseEncoding(raw);
  return RawParseUtils.Decode(enc, raw, msgB, raw.Length);
}
\end{C}
&
\begin{C}[style=CStyle,basicstyle=\tiny]
<@\textcolor{red}{\tb{\hl{static}}}@> string GetFullMessage() {
  byte[] raw = buffer;
  int msgB = RawParseUtils.TagMessage(raw, 0);
  if (msgB < 0 {
    return string.Empty;
  }
  Encoding enc = RawParseUtils.ParseEncoding(raw);
  return RawParseUtils.Decode(enc, raw, <@\textcolor{red}{\tb{\hl{MsgB}}}@>,raw.Length);
 }
\end{C}\\
\tiny{CodeBLEU$_\textnormal{before}$:77.09 } & \tiny{$\Delta_{drop}$: 10.42; \qquad CodeBLEU$_q$: 64.19} & \tiny{$\Delta_{drop}$: 21.93; \qquad CodeBLEU$_q$: 87.25} & \tiny{$\Delta_{drop}$: 22.8;\qquad CodeBLEU$_q$: 71.30}\\
\cline{1-4}
    \end{tabular*}
    \caption{Qualitative examples for the ablation study on  {\model}: Attack vulnerable tokens (VUL); with operator level constraints (VUL+OP), and with token level (VUL+OP+TOK) contraints on code translation task.}
    \label{tab:app_qual}
\end{table*}

\end{document}